\documentclass[10pt,twocolumn,letterpaper]{article}

\PassOptionsToPackage{table}{xcolor}

\usepackage{wacv}                       

\usepackage{graphicx}
\usepackage{amsmath,amssymb}
\usepackage{booktabs}
\usepackage{array}
\usepackage{multirow}
\usepackage{enumitem}   
\usepackage{tikz}       
\usetikzlibrary{arrows.meta,positioning,backgrounds,fit,shapes.geometric,calc}
\usepackage{pgfplots}   
\pgfplotsset{compat=1.18}

\usepackage{iftex}
\ifXeTeX
  \usepackage{fontspec}
\fi

\newcommand{\pipelineteaser}{%
  \vspace{0pt}\par\noindent
  \includegraphics[width=\textwidth]{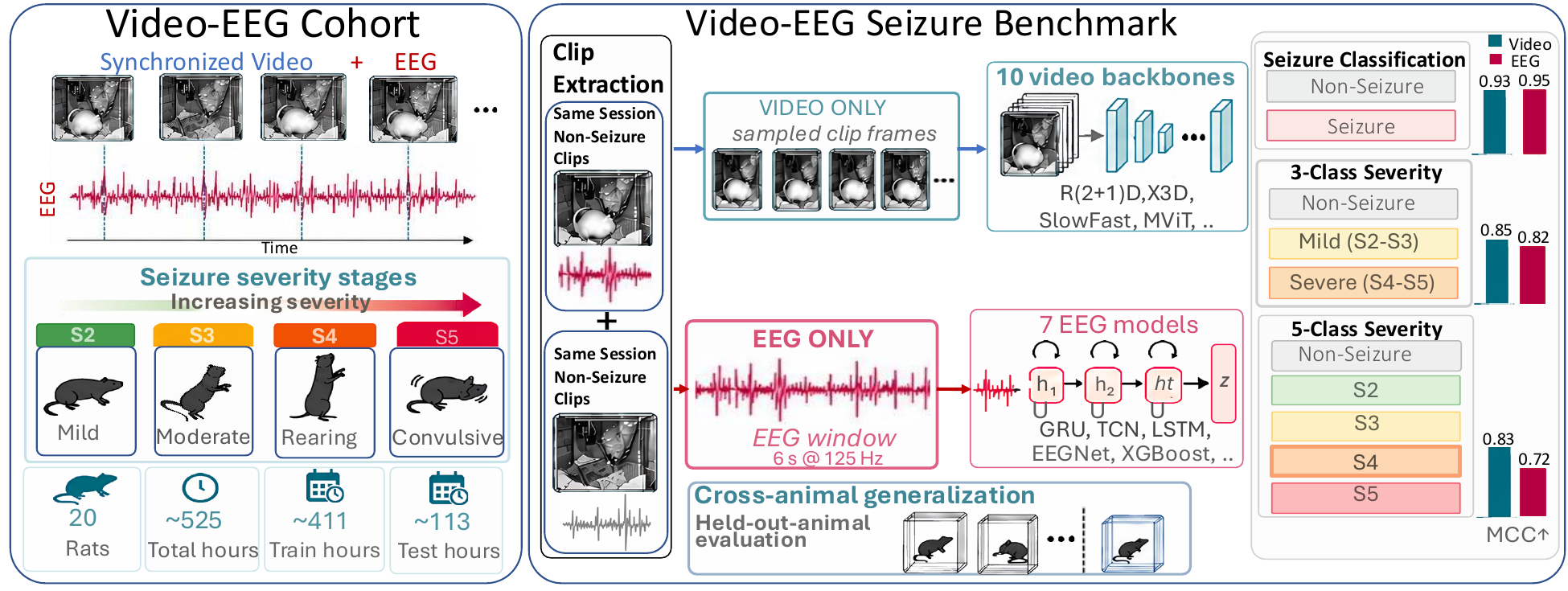}%
  \par\vspace{2pt}
  {\small Figure~1.\hspace{0.5em}Overview of the study. \emph{Left:} the cohort---
  synchronized video and EEG from 20 rats (corpus scale in
  App.~\ref{sec:dataextra:cohort}), each clip graded on a Racine-derived soman
  severity scale from Stage\,2 (mild) to Stage\,5 (convulsive). \emph{Right:} the
  benchmark. Expert-annotated seizure clips and same-session non-seizure clips feed
  two independent, unfused lanes---ten video backbones and seven EEG models---%
  compared head-to-head at 2-, 3- and 5-class granularity and under
  held-out-animal evaluation. Bars report MCC for detection and three- and five-class classification. The modalities are evaluated independently; no fusion results are shown.\par}%
  \vspace{3pt}%
}
\makeatletter
\g@addto@macro\@maketitle{\pipelineteaser}
\makeatother

\definecolor{wacvblue}{rgb}{0.21,0.49,0.74}
\usepackage[pagebackref,breaklinks,colorlinks,allcolors=wacvblue]{hyperref}

\def\wacvPaperID{2529} 
\def\confName{WACV}
\def\confYear{2027}

\title{Seeing Seizures: A Large-Scale Empirical Study of Video and EEG
Deep Learning for Automated Rodent Seizure Detection and Severity Grading}

\author{
Alloy Das$^{1}$\thanks{Corresponding author.}
\quad Shreyan Ganguly$^{1}$
\quad Nyzil Massey$^{2}$
\quad Hung Phi Doan$^{1}$\\
Zhanhong Jiang$^{1}$
\quad Thimmasettappa Thippeswamy$^{1}$
\quad Soumik Sarkar$^{1}$\\[1mm]
$^{1}$Iowa State University, Ames, IA, USA\\
$^{2}$Arkansas State University, Jonesboro, AR, USA\\[1mm]
{\tt\small \{alloydas, shreyang, hd1738, zhjiang, tswamy, soumiks\}@iastate.edu}\\
{\tt\small nmassey@astate.edu}
}

\begin{document}
\maketitle

\begin{abstract}
Manual annotation of synchronized video and EEG recordings is a bottleneck in preclinical seizure research. We present an empirical study of seizure detection and behavioral severity grading using 24,131 paired clips from 601 sessions across 20 rats in a soman-induced epilepsy model. Ten video architectures and seven single-channel EEG models are evaluated independently on binary detection and three- and five-class severity tasks. In the session-exclusive reference comparison, R(2+1)D-18 and GRU achieve detection macro-F1 scores of 0.968 and 0.974, respectively; for five-class grading, the corresponding scores are 0.657 and 0.560. Subject-exclusive evaluation reduces five-class macro-F1 to 0.475 for video and 0.404 for EEG, highlighting the difficulty of grading unseen animals. Temporal-sampling experiments demonstrate the value of multiple frames, while detection performance varies relatively little across the evaluated video architectures. A separate video-only binary-detection evaluation on RodEpil achieves a mean seizure-class F1 of 0.979 after uniform re-encoding to reduce identified class-correlated encoding cues. The study characterizes existing models for behavior-defined severity labels under the evaluated protocol, motivating further investigation of expert-assisted annotation workflows.  Our code and github page is available \href{https://alloydas.github.io/EEG-seizure-classification/}{here}.
\end{abstract}

\refstepcounter{figure}\label{fig:pipeline}

\section{Introduction}
\label{sec:intro}

Epilepsy affects roughly 1\% of the global population~\cite{liu2026}, driving animal research that relies heavily on chemically or electrically induced rodent models~\cite{bhatt2021}. These studies use synchronized video-EEG, pairing video of physical behavior with implanted brain sensors. Because electrical seizures often lack obvious movement and normal movement easily distorts EEG traces, using both tools together is an established reference approach~\cite{racine1972,lundt2015,bhatt2021}. Even though labs capture both data streams, automated seizure analysis focuses almost entirely on the electrical brain signals, using everything from basic frequency tracking to advanced deep learning~\cite{bergstrom2013,martin2023,ganguly2025}. As a result, the rich behavioral data captured on camera sits largely untouched by automated systems.

Properly assessing a seizure means doing more than just spotting an electrical brain event; it requires judging exactly how severe the physical behavior is. For our animals, we grade this severity on a custom five-point scale tailored for the soman model, since standard Racine scores do not directly map to chemically induced seizures~\cite{massey2025saracatinib,massey2026dualtargeting,racine1972,vanerum2019racine}. Physical symptoms escalate from mild tremors (Stage\,2) up to continuous, full-body convulsions (Stage\,5). Because Stage\,1 is incredibly rare, our labels focus strictly on Stages\,2 to 5. Because the represented severity categories are defined by observable motor behavior, video provides direct evidence for these labels. This motivates evaluating behavioral severity grading as a video-understanding task alongside EEG-based classification on the same annotated events.

This perspective connects seizure severity grading to a mature body of work
in video understanding. Modern action-recognition architectures are designed
to model temporally evolving visual patterns~\cite{carreira2017i3d,
tran2018r2plus1d,feichtenhofer2019slowfast}, while advances in computational
ethology provide complementary tools for quantifying animal
behavior~\cite{mathis2018deeplabcut,bohnslav2021deepethogram}. These methods
are natural candidates for recognizing severity-associated motor patterns such
as tremor, rearing, and generalized convulsions. Their application to
preclinical seizure analysis, however, remains limited: recent visual
benchmarks have focused primarily on seizure detection~\cite{rodepil2025},
while existing preclinical video-EEG studies predominantly address binary
detection~\cite{mullen2024,eegvfusion2026}. Consequently, it remains unclear
how effectively modern video models can recover behaviorally defined seizure
severity, and how their performance compares with EEG-based approaches on the
same synchronized events.
\begin{figure}[t]
\centering
\includegraphics[width=0.50\linewidth]{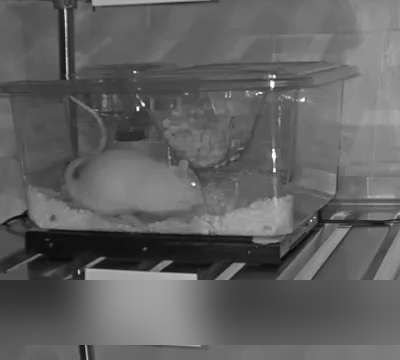}
\caption{Recording setup: each rat is recorded in its home cage using a fixed overhead camera and implanted EEG sensors. Cage cropping reduces variation in the field of view but does not remove all differences in camera geometry. Identifying details are masked.}
\label{fig:setup}
\end{figure}


\begin{figure}[t]
\centering
\includegraphics[width=\linewidth,  height=3.5cm]{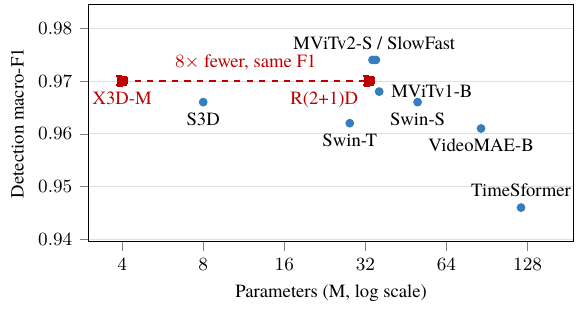}
\caption{Binary-detection macro-F1 versus video model size under the evaluated training budget. Scores range from 0.945 to 0.974. X3D-M (4M parameters) achieves 0.969, compared with 0.971 for R(2+1)D-18 (33M). This comparison concerns detection, not equivalence across severity tasks (Tables~\ref{tab:vidarch} and~\ref{tab:vidspec}).}
\label{fig:efficiency}
\end{figure}


We evaluate existing video and single-channel EEG models independently on matched events (Fig.~\ref{fig:pipeline}), examining task granularity, temporal sampling, model capacity, and generalization. Our target application is retrospective analysis of recorded seizures, where predictions could support expert review and behavioral scoring. This differs from forecasting an event before onset: these experiments classify extracted clips and do not assess pre-ictal prediction, preventive intervention, or reductions in annotation time.

The contribution is an empirical study rather than a new architecture or seizure-detection task. We make four contributions:
\begin{enumerate}
\item We characterize a synchronized video and EEG cohort containing 24,131 clips from 601 sessions across 20 rats, with behavioral labels spanning Stages~2 to 5.
\item We compare ten video architectures and seven single-channel EEG models on matched events and shared partitions at three levels of label granularity.
\item We examine architecture capacity and temporal sampling, and analyze class-specific errors and rare-stage performance.
\item We assess subject-exclusive generalization and video-only binary detection on RodEpil, documenting performance degradation and potential encoding shortcuts.
\end{enumerate}


\section{Related Work}
\label{sec:related}

Automated seizure tracking naturally splits based on the tools used and the subjects studied. We break past research down into three main areas: EEG-only, video-only, and mixed video and EEG systems.

\begin{figure}[t]
\centering
\includegraphics[width=\linewidth]{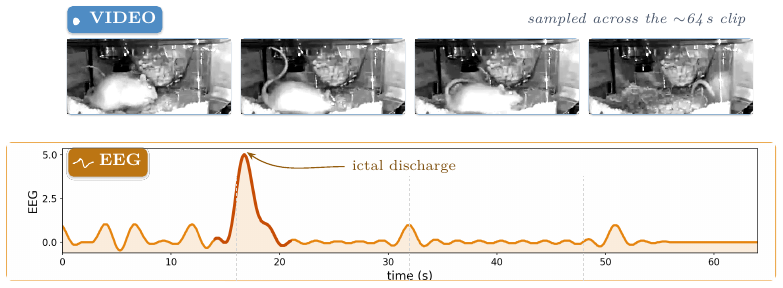}
\caption{A paired Stage\,3 seizure from rat R7. \emph{Top:} Four sample video frames. \emph{Bottom:} The corresponding single-channel EEG trace ($\sim$64\,s), with the active seizure period highlighted. Watching these streams side-by-side provides a complete view of both the physical symptoms and the electrical brain activity. Identifying markers have been masked.
}
\label{fig:paired}
\end{figure}

\noindent\textbf{EEG-based seizure analysis.}
EEG is still the standard tool for automated seizure tracking. Early animal systems relied on hand-picked wave features and basic machine learning~\cite{bergstrom2013}. Today, newer methods use deep neural networks, such as recurrent, convolutional, and attention models, to track signal changes over time~\cite{martin2023}. For chemically induced rodent seizures, deep recurrent networks like GRUs and LSTMs easily beat older tools like XGBoost and Random Forest~\cite{ganguly2025}. While these brain-wave models set a strong baseline for basic detection, they only watch electrical signals. They completely miss the physical movements that actually define a seizure's behavioral severity. This blind spot is exactly why this study compares them directly against video models on the exact same events.

\noindent\textbf{Video-based seizure and behavior analysis.}
Video seizure tracking ties closely to action recognition. Models like two-stream networks, 3D convolutions, and video transformers excel at mapping motion over time~\cite{simonyan2014twostream,carreira2017i3d,tran2018r2plus1d,feichtenhofer2019slowfast,bertasius2021timesformer,tong2022videomae}. In broader animal research, scientists already use similar tools to estimate poses~\cite{mathis2018deeplabcut,pereira2022sleap} and classify general behaviors~\cite{bohnslav2021deepethogram,hsu2021bsoid,sun2023mabe22}. For human epilepsy, clinical cameras mainly spot basic seizures and outward symptoms~\cite{ahmedt2018,karacsony2020seizure,vsvig2024}. However, animal video research remains fairly rare. Existing work mostly tracks specific induced seizures~\cite{sabnis2024}, general home-cage movement~\cite{jankovic2019homecage,epidetect2024}, or basic activity in mice~\cite{khan2024mice} and zebrafish~\cite{whytefagundes2025zebrafish}.

The closest matching study is Perlo \etal's \emph{RodEpil}~\cite{rodepil2025}. They built a video-only dataset with 13{,}053 clips from 19 rats and scored around 97\% F1 using a TimeSformer model~\cite{bertasius2021timesformer}. Yet, all of these prior animal studies stop at basic yes-or-no detection. Until now, using cameras to grade exact, multi-level seizure severity has never been systematically tested.

\noindent\textbf{Synchronized video and EEG analysis.}
Researchers rely heavily on paired video and EEG to assess seizures because physical behavior and EEG provide a complete picture (Fig.~\ref{fig:paired}). However, computer models combining both streams remain rare. Clinical trials have tested joint systems~\cite{aghaei_video_eeg,martini2021,cao2024,vepinet2024}, but animal studies usually stick to basic yes-or-no seizure detection using mixed sensors~\cite{mullen2024,eegvfusion2026}. Instead of mixing them right away, this study tests video and EEG separately on the exact same events. This allows a clean comparison from basic detection down to precise severity grading, setting the stage for future mixed models. A full literature review and breakdown are available in App.~\ref{sec:survey} and App.~\ref{sec:taxonomy}, Table~\ref{tab:gap}.

\section{Problem Setup}
\label{sec:problem}
We formalize the labels, task granularity, and modality-specific classifiers
used for the matched video and EEG evaluation in this section.
\subsection{Detection and severity grading}
\label{sec:problem:flat}
\noindent\textbf{Clips and labels.}
A session $r\in\mathcal{R}$ is one continuous synchronized video and EEG
recording from a single animal. From each session, we extract paired clips
$c=(\mathbf{V},\mathbf{E})$, including expert-annotated seizure events and
matched non-seizure intervals from the same session. The video clip is
$\mathbf{V}\in\mathbb{R}^{T\times H\times W\times 3}$, where $T$, $H$, and
$W$ denote the number of frames, frame height, and frame width, respectively.
The corresponding EEG clip is
$\mathbf{E}\in\mathbb{R}^{C\times L}$, where $C$ is the number of EEG
channels and $L$ is the number of temporal samples.

Each clip receives a label $y\in\{0,1,2,3,4\}$. The label $y=0$ denotes a
non-seizure clip, while $y=s-1$ represents Stage $s\in\{2,3,4,5\}$ on the
laboratory's soman-specific behavioral severity scale
\cite{massey2025saracatinib,massey2026dualtargeting}. Stage~1 is excluded
because it primarily reflects autonomic changes without reliable visible
motor manifestations and is extremely rare in the cohort. This scale is
adapted specifically to soman-induced seizures and should not be conflated
with the classical Racine scale
\cite{racine1972,vanerum2019racine}. Before pooling clips across animals,
each video is tightly cropped to the corresponding cage boundaries to account
for differences in camera geometry (Fig.~\ref{fig:setup}).

\noindent\textbf{Event boundaries.}
Experts mark a seizure's exact start and end at the first and last visible movements. Because EEG often shift before any physical signs appear, an extra 10\,s is added to both ends to catch the natural delay between the EEG and the video.

\begin{table}[t]
\centering
\caption{Reference R(2+1)D-18 video and single-channel GRU EEG results under session-exclusive evaluation (mean\,$\pm$\,std over five runs). P, R, and F1 are macro-averaged; MCC uses the full confusion matrix. Video has higher severity-classification macro-F1, whereas GRU has higher detection macro-F1. AUROC appears in App.~\ref{sec:extraresults:auroc}, Table~\ref{tab:aurocmain}. Best results are \textbf{bold}.}
\label{tab:results}

\setlength{\tabcolsep}{4pt}
\renewcommand{\arraystretch}{1.2}
\resizebox{\columnwidth}{!}{%
\begin{tabular}{@{}l l c c c c@{}}
\toprule
\textbf{Task} & \textbf{Modality} & \textbf{Prec.} & \textbf{Rec.} & \textbf{F1} & \textbf{MCC} \\
\midrule
\multirow{2}{*}{2-class (detection)} & Video & {0.968}\,{\scriptsize$\pm$0.002} & {0.968}\,{\scriptsize$\pm$0.002} & {0.968}\,{\scriptsize$\pm$0.002} & {0.936}\,{\scriptsize$\pm$0.004} \\
                                     & EEG   & \textbf{0.975}\,{\scriptsize$\pm$0.002} & \textbf{0.974}\,{\scriptsize$\pm$0.003} & \textbf{0.974}\,{\scriptsize$\pm$0.003} & \textbf{0.949}\,{\scriptsize$\pm$0.005} \\
\midrule
\multirow{2}{*}{3-class (severity)}  & Video & \textbf{0.775}\,{\scriptsize$\pm$0.016} & \textbf{0.788}\,{\scriptsize$\pm$0.009} & \textbf{0.780}\,{\scriptsize$\pm$0.007} & \textbf{0.853}\,{\scriptsize$\pm$0.010} \\
                                     & EEG   & {0.744}\,{\scriptsize$\pm$0.014} & {0.764}\,{\scriptsize$\pm$0.013} & {0.750}\,{\scriptsize$\pm$0.010} & {0.823}\,{\scriptsize$\pm$0.014} \\
\midrule
\multirow{2}{*}{5-class (severity)}  & Video & \textbf{0.689}\,{\scriptsize$\pm$0.024} & \textbf{0.643}\,{\scriptsize$\pm$0.006} & \textbf{0.657}\,{\scriptsize$\pm$0.009} & \textbf{0.831}\,{\scriptsize$\pm$0.006} \\
                                     & EEG   & {0.556}\,{\scriptsize$\pm$0.040} & {0.635}\,{\scriptsize$\pm$0.035} & {0.560}\,{\scriptsize$\pm$0.005} & {0.728}\,{\scriptsize$\pm$0.032} \\
\bottomrule
\end{tabular}}
\end{table}

\noindent\textbf{Task granularities.}
To track different levels of detail for the exact same events, three nested tasks map the main label $y$ using $g_k:\{0,\ldots,4\}\rightarrow\{0,\ldots,k-1\}$:
\begin{align}
g_2(y) &= \begin{cases}0,&y=0\\1,&y>0\end{cases} \label{eq:g2} \\[2mm]
g_3(y) &= \begin{cases}0,&y=0\\1,&y\in\{1,2\}\\2,&y\in\{3,4\}\end{cases} \label{eq:g3} \\[2mm]
g_5(y) &= y. \label{eq:g5}
\end{align}
Here, $g_2$ defines binary seizure detection; $g_3$ defines three-class
classification into non-seizure, lower-severity Stages~2 and 3, and
higher-severity Stages~4 and 5; and $g_5$ retains all four represented severity
stages as separate classes. The $g_3$ mapping forms two contiguous pairs of
the represented stages and serves as an intermediate-resolution grading
task.

\noindent\textbf{Modalities.}
For each task $\mathcal{T}_k$, where $k\in\{2,3,5\}$ is the number of
classes, we train separate video and EEG classifiers
$f_{\mathrm{V}}^{(k)}$ and $f_{\mathrm{E}}^{(k)}$. Let
$m\in\{\mathrm{V},\mathrm{E}\}$ denote the modality, with
$\mathbf{X}_{\mathrm{V}}=\mathbf{V}\in
\mathbb{R}^{T\times H\times W\times 3}$ and
$\mathbf{X}_{\mathrm{E}}=\mathbf{E}\in\mathbb{R}^{C\times L}$.
Each classifier outputs $k$ class probabilities, and its prediction is
\begin{equation}
\hat{y}_m^{(k)}
=
\arg\max_{j\in\{0,\ldots,k-1\}}
\left[f_m^{(k)}(\mathbf{X}_m)\right]_j,
\end{equation}
where $j$ indexes the output classes. The two modalities are trained and
evaluated independently on the same events; multimodal fusion is left for
future work (Sec.~\ref{sec:open}).

\noindent\textbf{Evaluation protocol.}
To prevent artificially high scores, several strict safeguards are in place. First, session-level splits keep entire recordings strictly in either training or testing for both tools. Second, inverse-frequency weighting balances out rare events. Third, macro-F1 serves as the main metric alongside MCC (App.~\ref{sec:metrics}); AUROC scores live in App.~\ref{sec:extraresults:auroc}. Finally, tests track individual results per rat and assess generalization to unseen animals using strict five-fold splits (Sec.~\ref{sec:crossmodal}).

\subsection{Conditional ordinal severity formulation}
\label{sec:problem:ordinal}

All reported experiments use the flat classifiers in Sec.~\ref{sec:problem:flat}. The following conditional ordinal formulation is a conceptual extension, not an implemented or evaluated component. Standard flat classifiers do not explicitly encode the ranked order of severity labels. A normal clip does not belong on this scale, while actual seizures follow a strict sequence
($\mathrm{S2}\prec\mathrm{S3}\prec\mathrm{S4}\prec\mathrm{S5}$). A basic classifier penalizes all mistakes equally, treating a minor S2-for-S3 mix-up exactly the same as a massive S2-for-S5 failure.

\noindent\textbf{Conditional factorization and ordinal head.}
In this proposed extension, seizure occurrence would be defined as $z=\mathbf{1}[y>0]$, splitting the prediction into two steps:
\begin{equation}
p(y\mid c)=
\begin{cases}
p(z=0\mid c), & y=0,\\[1mm]
p(z=1\mid c)\,p(s=s(y)\mid c,z=1), & y>0.
\end{cases}
\label{eq:conditional}
\end{equation}
For active seizures, severity could be represented using cumulative probabilities:
\begin{equation}
q_j(c)=p(s\succ\mathrm{S}_j\mid c,z=1),\qquad j\in\{2,3,4\},
\end{equation}
subject to the monotonicity constraint
$q_2(c)\ge q_3(c)\ge q_4(c)$.
A future implementation could derive the predicted stage from thresholded cumulative probabilities.

\noindent\textbf{Ordinal evaluation.}
Macro-F1 misses how far a wrong guess falls from the true severity stage. Future evaluations could measure the average absolute difference between predicted and reference stages,
$\mathrm{MAE}_{\mathrm{stage}}
=\frac{1}{N}\sum_{i=1}^{N}\left|\hat{s}_i-s_i\right|$,
and the frequency of errors spanning two or more stages,
$\frac{1}{N}\sum_{i=1}^{N}
\mathbf{1}\left[\left|\hat{s}_i-s_i\right|\ge 2\right].$
Normal clips do not fit this ranked scale, so these metrics only apply to actual seizures. Future ranking tests should use the exact same unseen-animal splits from Sec.~\ref{sec:crossmodal}.

\section{Dataset}
\label{sec:data}

\begin{table*}[t]
\centering
\caption{We tested ten video models using the exact same data splits and a strict 12-epoch training limit. Scores show the mean\,$\pm$\,std over five test runs (SlowFast detection uses $n{=}4$). Basic detection scores barely changed, shifting by only 0.03 across all the models. You can find the AUROC scores and exact model setups in App.~\ref{sec:extraresults:auroc} and App.~\ref{sec:arch:video}. We \textbf{bolded} the best results and \underline{underlined} the second-best.
}
\label{tab:vidarch}

\setlength{\tabcolsep}{5pt}
\renewcommand{\arraystretch}{1.12}
\resizebox{\textwidth}{!}{%
\begin{tabular}{@{}l cccc cccc cccc@{}}
\toprule
 & \multicolumn{4}{c}{\textbf{Detection (2-class)}} & \multicolumn{4}{c}{\textbf{3-class grading}} & \multicolumn{4}{c}{\textbf{5-class grading}} \\
\cmidrule(lr){2-5}\cmidrule(lr){6-9}\cmidrule(lr){10-13}
\textbf{Backbone} & \textbf{P} & \textbf{R} & \textbf{F1} & \textbf{MCC} & \textbf{P} & \textbf{R} & \textbf{F1} & \textbf{MCC} & \textbf{P} & \textbf{R} & \textbf{F1} & \textbf{MCC} \\
\midrule
MViTv2-S~\cite{li2022mvitv2} & $\underline{0.972}_{\pm0.001}$ & $\underline{0.972}_{\pm0.001}$ & $\underline{0.972}_{\pm0.001}$ & $\underline{0.944}_{\pm0.003}$ & $0.780_{\pm0.014}$ & $0.800_{\pm0.027}$ & $0.786_{\pm0.005}$ & $0.856_{\pm0.017}$ & $0.694_{\pm0.032}$ & $\mathbf{0.656}_{\pm0.030}$ & $0.660_{\pm0.016}$ & $0.832_{\pm0.007}$ \\
SlowFast-R50~\cite{feichtenhofer2019slowfast} & $\mathbf{0.974}_{\pm0.002}$ & $\mathbf{0.974}_{\pm0.002}$ & $\mathbf{0.974}_{\pm0.002}$ & $\mathbf{0.948}_{\pm0.004}$ & $\underline{0.781}_{\pm0.018}$ & $\underline{0.804}_{\pm0.013}$ & $\underline{0.789}_{\pm0.009}$ & $\underline{0.859}_{\pm0.012}$ & $\underline{0.718}_{\pm0.024}$ & $\underline{0.655}_{\pm0.014}$ & $\mathbf{0.675}_{\pm0.008}$ & $\mathbf{0.851}_{\pm0.003}$ \\
R(2+1)D-18~\cite{tran2018r2plus1d} & $0.971_{\pm0.001}$ & $0.971_{\pm0.001}$ & $0.971_{\pm0.001}$ & $0.942_{\pm0.002}$ & $0.780_{\pm0.007}$ & $0.787_{\pm0.007}$ & $0.783_{\pm0.002}$ & $\mathbf{0.860}_{\pm0.003}$ & $\mathbf{0.734}_{\pm0.027}$ & $0.643_{\pm0.011}$ & $\underline{0.667}_{\pm0.010}$ & $\underline{0.841}_{\pm0.007}$ \\
X3D-M~\cite{feichtenhofer2020x3d} & $0.969_{\pm0.002}$ & $0.969_{\pm0.002}$ & $0.969_{\pm0.002}$ & $0.937_{\pm0.003}$ & $\mathbf{0.782}_{\pm0.008}$ & $\mathbf{0.810}_{\pm0.015}$ & $\mathbf{0.794}_{\pm0.005}$ & $0.856_{\pm0.008}$ & $0.568_{\pm0.004}$ & $0.616_{\pm0.013}$ & $0.586_{\pm0.003}$ & $0.824_{\pm0.007}$ \\
MViTv1-B~\cite{fan2021mvit} & $0.969_{\pm0.002}$ & $0.969_{\pm0.002}$ & $0.969_{\pm0.002}$ & $0.938_{\pm0.003}$ & $0.771_{\pm0.018}$ & $0.794_{\pm0.016}$ & $0.779_{\pm0.011}$ & $0.849_{\pm0.016}$ & $0.687_{\pm0.022}$ & $0.643_{\pm0.011}$ & $0.653_{\pm0.008}$ & $0.836_{\pm0.006}$ \\
Video Swin-S~\cite{liu2022videoswin} & $0.966_{\pm0.003}$ & $0.966_{\pm0.003}$ & $0.966_{\pm0.003}$ & $0.932_{\pm0.005}$ & $0.767_{\pm0.013}$ & $0.780_{\pm0.022}$ & $0.771_{\pm0.009}$ & $0.834_{\pm0.016}$ & $0.716_{\pm0.032}$ & $0.634_{\pm0.017}$ & $0.645_{\pm0.011}$ & $0.818_{\pm0.012}$ \\
S3D~\cite{xie2018s3d} & $0.967_{\pm0.002}$ & $0.967_{\pm0.002}$ & $0.967_{\pm0.002}$ & $0.933_{\pm0.005}$ & $0.756_{\pm0.014}$ & $0.782_{\pm0.027}$ & $0.764_{\pm0.004}$ & $0.840_{\pm0.015}$ & $0.644_{\pm0.015}$ & $0.639_{\pm0.021}$ & $0.636_{\pm0.009}$ & $0.813_{\pm0.015}$ \\
Video Swin-T~\cite{liu2022videoswin} & $0.963_{\pm0.001}$ & $0.962_{\pm0.001}$ & $0.963_{\pm0.001}$ & $0.925_{\pm0.002}$ & $0.761_{\pm0.014}$ & $0.776_{\pm0.020}$ & $0.766_{\pm0.008}$ & $0.829_{\pm0.006}$ & $0.660_{\pm0.022}$ & $0.639_{\pm0.009}$ & $0.632_{\pm0.002}$ & $0.796_{\pm0.006}$ \\
VideoMAE-B~\cite{tong2022videomae} & $0.958_{\pm0.004}$ & $0.958_{\pm0.004}$ & $0.958_{\pm0.004}$ & $0.915_{\pm0.007}$ & $0.729_{\pm0.033}$ & $0.736_{\pm0.042}$ & $0.729_{\pm0.030}$ & $0.771_{\pm0.063}$ & $0.704_{\pm0.018}$ & $0.623_{\pm0.025}$ & $0.636_{\pm0.017}$ & $0.802_{\pm0.013}$ \\
TimeSformer~\cite{bertasius2021timesformer} & $0.945_{\pm0.001}$ & $0.944_{\pm0.001}$ & $0.945_{\pm0.001}$ & $0.889_{\pm0.002}$ & $0.722_{\pm0.017}$ & $0.761_{\pm0.017}$ & $0.735_{\pm0.013}$ & $0.778_{\pm0.019}$ & $0.650_{\pm0.014}$ & $0.594_{\pm0.014}$ & $0.601_{\pm0.009}$ & $0.757_{\pm0.028}$ \\
\bottomrule
\end{tabular}}
\end{table*}

We analyzed paired video and EEG recordings from twenty rats exposed to soman. The dataset contains 24{,}131 clips across 601 sessions. Following standard rules~\cite{massey2025saracatinib,massey2026dualtargeting}, we recorded each rat in its home cage using a fixed overhead camera and implanted EEG sensors. The main recording system stamped the time on both streams to synchronize their timestamps. We tracked natural seizures that happened 10 to 18 weeks after the initial exposure, right in the middle of the chronic epilepsy phase. We recorded the raw EEG at 1\,kHz and lowered the rate to 125\,Hz. Finally, we cropped the videos tightly around each cage to reduce differences in the field of view (Fig.~\ref{fig:setup}).

\noindent\textbf{Annotation and labels.}
Researchers reviewed synchronized video and EEG recordings using NeuroScore (DSI), marked event boundaries, and assigned behavioral severity labels~\cite{massey2025saracatinib,massey2026dualtargeting}. NeuroScore supported this review workflow; labels were obtained through researcher review rather than accepted as unverified software outputs. EEG alerts were checked against video to identify false alarms. The reference labels are therefore not modality-independent. Formal inter-rater agreement is not reported, leaving annotation reproducibility across independent raters unquantified (Sec.~\ref{sec:discussion}).

We grade seizure severity on a 5-level scale~\cite{racine1972,massey2025saracatinib,massey2026dualtargeting}. This is very different from the classic Racine scale~\cite{vanerum2019racine}. Stage\,1 only involves internal body changes. Stages\,2 through 5 track worse and worse physical shaking, ending in full-body convulsions. We drop Stage\,1 because we only caught 6 clips of it. The working data has 1{,}435 Stage\,2, 9{,}256 Stage\,3, 1{,}050 Stage\,4, and 193 Stage\,5 clips. When we add 12{,}197 normal baseline clips, the final total is 24{,}131 clips. The data is heavily lopsided across both the seizure stages and the individual rats. To handle this, we give rare events extra weight, carefully average our scores, track results per rat, and test the models on brand new animals. You can find all the strict rules for ethics, privacy, and data handling in App.~\ref{sec:dataextra} (App.~\ref{sec:dataextra:governance}). While we are not able to release the complete raw data at this time due to certain institutional restrictions, a future release through a scientific repository is planned, subject to institutional approval. The complete raw cohort is not currently publicly available. For reproducibility, we provide the data collection protocol in the App.~\ref{sec:dataextra} (App.~\ref{sec:dataextra:acq}).

\section{Experiments and Discussion}
\label{sec:application}

We test standard video and EEG models on our group of twenty rats (Sec.~\ref{sec:data}). We want to see how well they handle basic seizure detection and exact severity grading. We also check how they perform on brand new animals and on outside datasets. Unless we say otherwise, the video setup uses a R(2+1)D-18 model~\cite{tran2018r2plus1d,kay2017kinetics} watching 16 frames at a $112\times112$ resolution. The EEG setup uses the GRU model built by Ganguly~\etal~\cite{ganguly2025}. Both tools face the exact same test splits and use the same math rules to balance out rare events. Table~\ref{tab:results} shows the average scores across five different test runs. You can find all the strict details about the model designs and training steps in Apps.~\ref{sec:arch:video}~\ref{sec:arch:train}.

\begin{table}[!ht]
\centering
\caption{Temporal-sampling ablation for R(2+1)D (seed 42). Single-frame performance is poor, while performance peaks around 32 frames. Using 64 frames gives no detection gain and lowers 5-class F1. AUROC results are reported in App.~\ref{sec:extraresults:auroc}, Table~\ref{tab:aurocsweep}.}
\label{tab:frames}

\setlength{\tabcolsep}{3.5pt}
\renewcommand{\arraystretch}{1.12}
\resizebox{\columnwidth}{!}{%
\begin{tabular}{@{}c cc cc cc@{}}
\toprule
 & \multicolumn{2}{c}{\textbf{Detection}} & \multicolumn{2}{c}{\textbf{3-class}} & \multicolumn{2}{c}{\textbf{5-class}} \\
\cmidrule(lr){2-3}\cmidrule(lr){4-5}\cmidrule(lr){6-7}
\textbf{Frames} & \textbf{F1} & \textbf{MCC} & \textbf{F1} & \textbf{MCC} & \textbf{F1} & \textbf{MCC} \\
\midrule
1  & 0.569 & 0.149 & 0.417 & 0.152 & 0.314 & 0.147 \\
4  & 0.940 & 0.881 & 0.739 & 0.760 & 0.612 & 0.741 \\
8  & 0.959 & 0.918 & 0.763 & 0.838 & 0.627 & 0.789 \\
16 & 0.970 & 0.941 & 0.780 & \underline{0.857} & \underline{0.662} & \underline{0.830} \\
32 & \textbf{0.972} & \textbf{0.943} & \textbf{0.793} & \textbf{0.858} & \textbf{0.667} & \textbf{0.835} \\
64 & \underline{0.971} & \underline{0.941} & \underline{0.785} & 0.848 & 0.632 & 0.807 \\
\bottomrule
\end{tabular}}
\end{table}

\subsection{Video capacity and input requirements}
\label{sec:rq1}

\noindent\textbf{Detection baseline.}
Our baseline R(2+1)D model scores $0.968\pm0.002$ macro-F1 and $0.936\pm0.004$ MCC (Table~\ref{tab:results}). We purposely paired seizure and normal clips from the exact same session. This keeps the animal, camera angle, and lighting identical. This reduces opportunities to distinguish classes using session-specific background cues (App.~\ref{sec:dataextra:stages}, Fig.~\ref{fig:detexample}). The model also stays highly accurate for the animals that suffer the most seizures, achieving macro-F1 scores of 0.974 on rat R1 and 0.906 on rat R7.

\begin{table*}[t]
\centering
\caption{Seven EEG baselines use the same 6\,s windows, 3\,s stride, 125\,Hz sampling, and log-mean aggregation (App.~\ref{sec:eegsweep}). TCN performs best for detection and 3-class grading, while GRU and TCN are similar at 5 classes. Two LSTM runs failed to converge, so its 3- and 5-class results use $n{=}4$. Best results are in \textbf{bold}; second-best are \underline{underlined}.}
\label{tab:eegarch}

\setlength{\tabcolsep}{3pt}
\renewcommand{\arraystretch}{1.15}
\resizebox{\textwidth}{!}{%
\begin{tabular}{@{}l cccc cccc cccc@{}}
\toprule
 & \multicolumn{4}{c}{\textbf{Detection (2-class)}} & \multicolumn{4}{c}{\textbf{3-class}} & \multicolumn{4}{c}{\textbf{5-class}} \\
\cmidrule(lr){2-5}\cmidrule(lr){6-9}\cmidrule(lr){10-13}
\textbf{EEG backbone} & \textbf{P} & \textbf{R} & \textbf{F1} & \textbf{MCC} & \textbf{P} & \textbf{R} & \textbf{F1} & \textbf{MCC} & \textbf{P} & \textbf{R} & \textbf{F1} & \textbf{MCC} \\
\midrule
\multicolumn{13}{@{}l}{\itshape Deep}\\
GRU~\cite{ganguly2025} & \underline{0.975}\,{\scriptsize$\pm$0.003} & \underline{0.974}\,{\scriptsize$\pm$0.003} & \underline{0.974}\,{\scriptsize$\pm$0.003} & \underline{0.949}\,{\scriptsize$\pm$0.005} & \underline{0.744}\,{\scriptsize$\pm$0.014} & 0.764\,{\scriptsize$\pm$0.013} & \underline{0.750}\,{\scriptsize$\pm$0.010} & \underline{0.823}\,{\scriptsize$\pm$0.014} & \textbf{0.556}\,{\scriptsize$\pm$0.040} & \underline{0.635}\,{\scriptsize$\pm$0.035} & \underline{0.560}\,{\scriptsize$\pm$0.005} & \textbf{0.728}\,{\scriptsize$\pm$0.032} \\
LSTM~\cite{hochreiter1997lstm} & 0.972\,{\scriptsize$\pm$0.008} & 0.971\,{\scriptsize$\pm$0.007} & 0.971\,{\scriptsize$\pm$0.008} & 0.942\,{\scriptsize$\pm$0.015} & 0.696\,{\scriptsize$\pm$0.041} & 0.741\,{\scriptsize$\pm$0.067} & 0.698\,{\scriptsize$\pm$0.053} & 0.743\,{\scriptsize$\pm$0.059} & 0.476\,{\scriptsize$\pm$0.042} & 0.555\,{\scriptsize$\pm$0.100} & 0.468\,{\scriptsize$\pm$0.052} & 0.598\,{\scriptsize$\pm$0.059} \\
EEGNet~\cite{lawhern2018eegnet} & 0.926\,{\scriptsize$\pm$0.011} & 0.922\,{\scriptsize$\pm$0.013} & 0.923\,{\scriptsize$\pm$0.012} & 0.848\,{\scriptsize$\pm$0.023} & 0.646\,{\scriptsize$\pm$0.010} & 0.676\,{\scriptsize$\pm$0.018} & 0.620\,{\scriptsize$\pm$0.015} & 0.626\,{\scriptsize$\pm$0.031} & 0.448\,{\scriptsize$\pm$0.040} & 0.488\,{\scriptsize$\pm$0.025} & 0.409\,{\scriptsize$\pm$0.039} & 0.499\,{\scriptsize$\pm$0.037} \\
EEG Conformer~\cite{song2022conformer} & 0.972\,{\scriptsize$\pm$0.001} & 0.972\,{\scriptsize$\pm$0.001} & 0.972\,{\scriptsize$\pm$0.001} & 0.943\,{\scriptsize$\pm$0.001} & 0.719\,{\scriptsize$\pm$0.009} & \underline{0.784}\,{\scriptsize$\pm$0.013} & 0.729\,{\scriptsize$\pm$0.010} & 0.780\,{\scriptsize$\pm$0.017} & 0.512\,{\scriptsize$\pm$0.004} & 0.624\,{\scriptsize$\pm$0.032} & 0.523\,{\scriptsize$\pm$0.005} & 0.683\,{\scriptsize$\pm$0.008} \\
TCN~\cite{bai2018tcn} & \textbf{0.979}\,{\scriptsize$\pm$0.001} & \textbf{0.978}\,{\scriptsize$\pm$0.001} & \textbf{0.978}\,{\scriptsize$\pm$0.001} & \textbf{0.957}\,{\scriptsize$\pm$0.002} & \textbf{0.758}\,{\scriptsize$\pm$0.014} & \textbf{0.786}\,{\scriptsize$\pm$0.008} & \textbf{0.767}\,{\scriptsize$\pm$0.007} & \textbf{0.839}\,{\scriptsize$\pm$0.015} & \underline{0.537}\,{\scriptsize$\pm$0.008} & \textbf{0.654}\,{\scriptsize$\pm$0.013} & \textbf{0.563}\,{\scriptsize$\pm$0.012} & \underline{0.728}\,{\scriptsize$\pm$0.021} \\
\midrule
\multicolumn{13}{@{}l}{\itshape Classical (band-power $+$ statistical features)}\\
Random Forest~\cite{breiman2001rf} & 0.885\,{\scriptsize$\pm$0.001} & 0.852\,{\scriptsize$\pm$0.001} & 0.852\,{\scriptsize$\pm$0.001} & 0.737\,{\scriptsize$\pm$0.002} & 0.541\,{\scriptsize$\pm$0.001} & 0.542\,{\scriptsize$\pm$0.001} & 0.529\,{\scriptsize$\pm$0.001} & 0.613\,{\scriptsize$\pm$0.002} & 0.477\,{\scriptsize$\pm$0.001} & 0.328\,{\scriptsize$\pm$0.001} & 0.332\,{\scriptsize$\pm$0.002} & 0.503\,{\scriptsize$\pm$0.002} \\
XGBoost~\cite{chen2016xgboost} & 0.922\,{\scriptsize$\pm$0.000} & 0.913\,{\scriptsize$\pm$0.000} & 0.914\,{\scriptsize$\pm$0.000} & 0.835\,{\scriptsize$\pm$0.000} & 0.657\,{\scriptsize$\pm$0.000} & 0.684\,{\scriptsize$\pm$0.000} & 0.642\,{\scriptsize$\pm$0.000} & 0.623\,{\scriptsize$\pm$0.000} & 0.479\,{\scriptsize$\pm$0.000} & 0.530\,{\scriptsize$\pm$0.000} & 0.466\,{\scriptsize$\pm$0.000} & 0.529\,{\scriptsize$\pm$0.000} \\
\bottomrule
\end{tabular}}
\end{table*}

\noindent\textbf{Backbone capacity.}
Under the evaluated training budget, binary-detection macro-F1 ranges from 0.945 to 0.974 across ten video architectures (Table~\ref{tab:vidarch}). X3D-M (4M parameters) achieves 0.969, compared with 0.971 for R(2+1)D-18 (33M; Fig.~\ref{fig:efficiency}). Their five-class macro-F1 scores differ more substantially (0.586 and 0.667), so similar detection performance does not imply equivalent severity grading. The separate data-size experiment evaluates only R(2+1)D detection: using ten training animals yields 0.957 macro-F1 versus 0.970 with twenty. It does not establish data saturation for larger architectures or other tasks (App.~\ref{sec:extraresults:dataeff}).

\noindent\textbf{Temporal and spatial evidence.}
Tracking motion over time is incredibly important. A model looking at just a single still frame does very poorly. Its macro-F1 scores are 0.569, 0.417, and 0.314 across the 2, 3, and 5-level tests. Using 16 frames works much better, bumping those scores up to 0.970, 0.780, and 0.662 (Table~\ref{tab:frames}). Performance peaks around 32 frames. Pushing it up to 64 frames does not improve basic detection at all and actually hurts the 5-level grading, dropping it to 0.632. On the flip side, making the video sharper does not help. Increasing the resolution from 112 to 224 or 448 pixels gives zero benefit (App.~\ref{sec:extraresults:ablations}). This makes sense because the original lab cameras only shoot at around 300 to 480 pixels anyway. Because of this, we set our standard setup to 16 frames at a $112\times112$ resolution.

\begin{figure}[t]
\centering
\includegraphics[width=\linewidth, height=3cm, trim=0 0 0 42, clip]{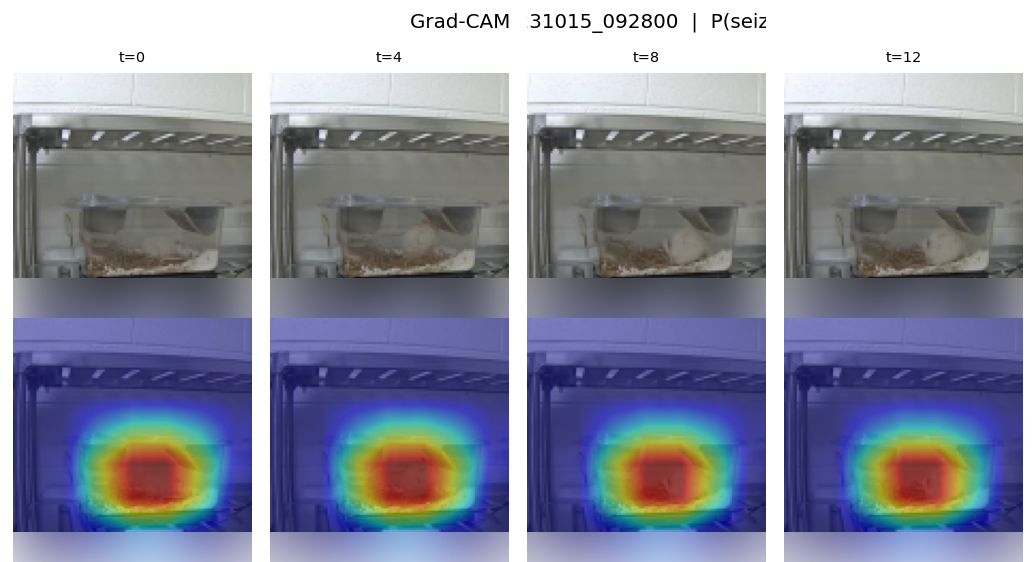}\\[1pt]
{\footnotesize (a) Seizure clip (Stage\,4): activation concentrates on the animal}\\[4pt]
\includegraphics[width=\linewidth, height=3cm, trim=0 0 0 42, clip]{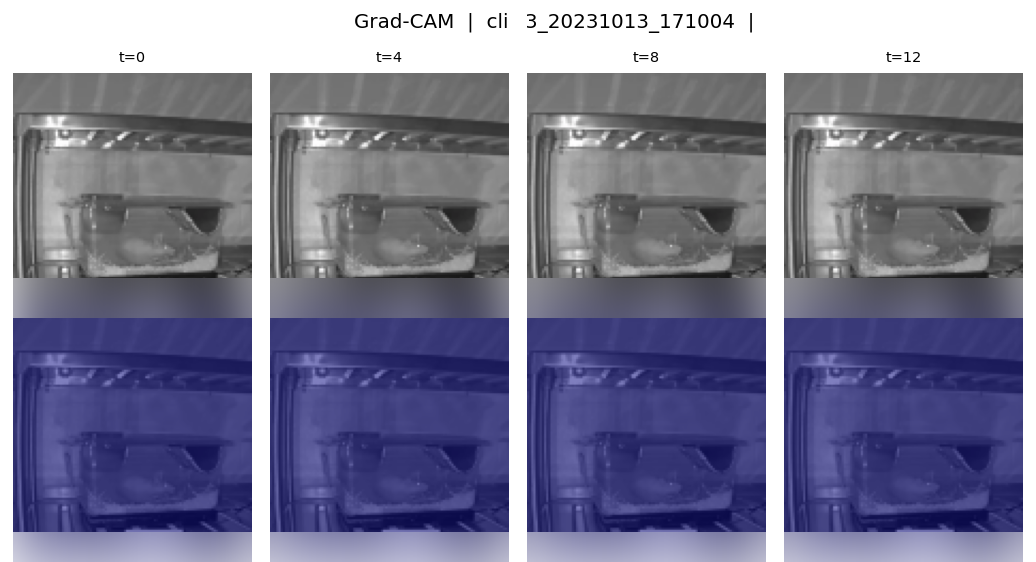}\\[1pt]
{\footnotesize (b) Non-seizure clip: activation is diffuse, with no focus on the animal}
\caption{Grad-CAM~\cite{selvaraju2017gradcam} for R(2+1)D-18. Four of eight frames are shown, with input frames above activation maps. Activations concentrate on the animal in the Stage~4 example and are more diffuse in the non-seizure example. These observations do not establish causal reliance on behavior or exclude shortcuts. Full sequences appear in App.~\ref{sec:qualitative:gradcam}, Fig.~\ref{fig:gradcamfull}. Identifying tags are blurred.}
\label{fig:gradcam}
\end{figure}

\noindent\textbf{Shortcut check.}
Same-session pairing and cage cropping reduce some potential shortcuts. The single-frame ablation supports the importance of temporal evidence in this rodent setup; it does not establish that static images are uninformative in other settings or species. Grad-CAM highlights the animal in the illustrated seizure clip (Fig.~\ref{fig:gradcam}), but these observations do not exclude all shortcuts. Animal-masking, background-only, and temporal-shuffling controls would provide additional evidence (App.~\ref{sec:qualitative:shortcuts}).

\subsection{Cross-modal comparison and generalization}
\label{sec:crossmodal}

\noindent\textbf{EEG baseline strength.}
We evaluate GRU, LSTM~\cite{hochreiter1997lstm}, EEGNet~\cite{lawhern2018eegnet}, EEG Conformer~\cite{song2022conformer}, TCN~\cite{bai2018tcn}, Random Forest~\cite{breiman2001rf}, and XGBoost~\cite{chen2016xgboost} using the same single EEG channel and preprocessing. The selected configuration uses 6-second windows at 125\,Hz (App.~\ref{sec:eegsweep}). TCN achieves the highest detection and three-class macro-F1 among these EEG models (0.978 and 0.767), and five-class macro-F1 of 0.563 versus 0.560 for GRU. The reference video model achieves 0.968, 0.780, and 0.657, respectively. These comparisons are conditional on the evaluated configurations; they do not establish performance relative to multichannel EEG.

\noindent\textbf{Metric robustness.}
For the session-exclusive reference models, the video-minus-GRU macro-F1 difference is $-0.006$, $+0.030$, and $+0.097$ for two-, three-, and five-class classification. The MCC differences are $-0.013$, $+0.030$, and $+0.103$. Both metrics favor GRU for detection and video for severity classification in this comparison (Table~\ref{tab:results}). AUROC is reported separately in App.~\ref{sec:extraresults:auroc}.

\begin{figure}[t]
\centering
\includegraphics[width=0.75\linewidth]{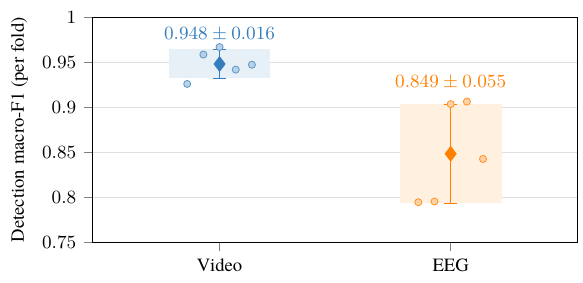}
\caption{Per-fold binary-detection macro-F1 for unseen animals, as displayed in the plot: video $0.948\pm0.016$ and EEG $0.849\pm0.055$. The plotted video results have a higher mean and lower fold-to-fold variability than the plotted EEG results.}
\label{fig:stability}
\end{figure}

\noindent\textbf{Unseen animals.}
Subject-exclusive evaluation reduces severity-classification performance. Table~\ref{tab:subjectwise} reports three-class macro-F1 of 0.720 for video and 0.656 for EEG, and five-class macro-F1 of 0.475 and 0.404, compared with session-exclusive values of 0.657 and 0.560. The five-class difference therefore decreases from 0.097 to 0.071. Figure~\ref{fig:stability} displays per-fold detection macro-F1 of $0.948\pm0.016$ for video and $0.849\pm0.055$ for EEG.

\begin{table}[tb]
\centering
\caption{Testing new sessions versus new animals. Testing unseen animals keeps basic detection high but hurts exact grading. Video's 5-class lead shrinks from 0.097 to 0.071. Scores show the mean\,$\pm$\,std over five tests.}
\label{tab:subjectwise}

\setlength{\tabcolsep}{3pt}
\renewcommand{\arraystretch}{1.2}
\resizebox{\columnwidth}{!}{%
\begin{tabular}{@{}l l cc cc@{}}
\toprule
 & & \multicolumn{2}{c}{\textbf{Session-excl.}} & \multicolumn{2}{c}{\textbf{Subject-exclusive (5-fold)}} \\
\cmidrule(lr){3-4}\cmidrule(lr){5-6}
\textbf{Task} & \textbf{Mod.} & \textbf{Macro-F1} & \textbf{MCC} & \textbf{Macro-F1} & \textbf{MCC} \\
\midrule
\multirow{2}{*}{Detection} & Video & 0.968 & 0.936 & 0.948\,{\scriptsize$\pm$0.016} & 0.896\,{\scriptsize$\pm$0.031} \\
                           & EEG   & \textbf{0.974} & \textbf{0.949} & \textbf{0.972}\,{\scriptsize$\pm$0.015} & \textbf{0.944}\,{\scriptsize$\pm$0.030} \\
\midrule
\multirow{2}{*}{3-class}   & Video & \textbf{0.780} & \textbf{0.853} & \textbf{0.720}\,{\scriptsize$\pm$0.028} & 0.761\,{\scriptsize$\pm$0.029} \\
                           & EEG   & 0.750 & 0.823 & 0.656\,{\scriptsize$\pm$0.022} & \textbf{0.793}\,{\scriptsize$\pm$0.046} \\
\midrule
\multirow{2}{*}{5-class}   & Video & \textbf{0.657} & \textbf{0.831} & \textbf{0.475}\,{\scriptsize$\pm$0.024} & \textbf{0.676}\,{\scriptsize$\pm$0.041} \\
                           & EEG   & 0.560 & 0.728 & 0.404\,{\scriptsize$\pm$0.042} & 0.657\,{\scriptsize$\pm$0.107} \\
\bottomrule
\end{tabular}}
\end{table}

\begin{figure}[t]
\centering
\includegraphics[width=0.8\linewidth]{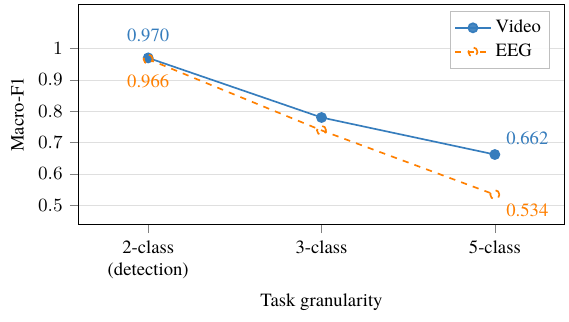}
\caption{Macro-F1 across task granularities for the results displayed in this plot. Video decreases from 0.970 for binary detection to 0.662 for five-class classification; EEG decreases from 0.966 to 0.534. Both modalities perform less well as the label set becomes more detailed. The plotted values differ from the run-averaged values in Table~\ref{tab:results}.}
\label{fig:degrade}
\end{figure}

\begin{figure}[t]
\centering
\includegraphics[width=0.8\linewidth]{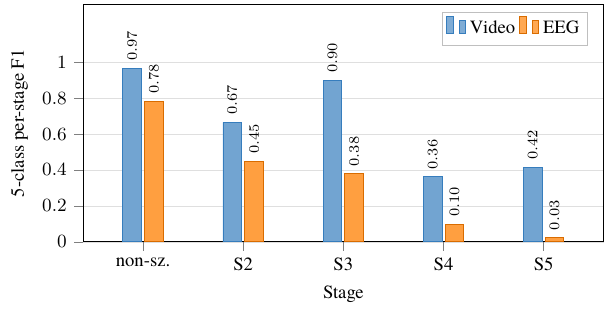}
\caption{Per-class F1 for the five-class results displayed in this plot. Video/EEG values are 0.97/0.78 for non-seizure, 0.67/0.45 for Stage~2, 0.90/0.38 for Stage~3, 0.36/0.10 for Stage~4, and 0.42/0.03 for Stage~5. Rare severe stages remain difficult for both modalities.}
\label{fig:perclassbar}
\end{figure}

\noindent\textbf{Granularity and error complementarity.}
The session-exclusive averages in Table~\ref{tab:results} decrease from 0.968 to 0.780 to 0.657 macro-F1 for video, and from 0.974 to 0.750 to 0.560 for GRU as label granularity increases. Figure~\ref{fig:degrade} also shows declining macro-F1 with increasing granularity. Rare severe stages remain difficult: Figure~\ref{fig:perclassbar} shows video F1 of 0.36 and 0.42 for Stages~4 and~5, versus EEG F1 of 0.10 and 0.03. Aggregate performance should not be interpreted as uniformly reliable severity grading.

The two tools also make different types of mistakes. The diagnostic recall analysis reports higher EEG recall for Stage\,2 (0.863 versus 0.650 for video). But EEG completely misses 10.1\% of the strong Stage\,3 seizures, calling them normal. Video only misses 2.3\% of these. We looked at 2{,}830 clips tested by both models. The video was the only one right 14.2\% of the time. The EEG was the only one right 4.9\% of the time (App.~\ref{sec:extraresults:failure}, Table~\ref{tab:paired}). These differing errors motivate evaluating fusion; they do not establish that fusion would improve performance.

\noindent\textbf{External validation.}
We separately evaluate R(2+1)D on RodEpil~\cite{rodepil2025} for video-only binary seizure detection. The original files contain class-correlated encoding differences: non-seizure clips have 301 frames at 30\,fps, whereas seizure clips have 291 frames at 29\,fps, with additional framing differences. We uniformly re-encode videos to 300 frames at 30\,fps, letterbox them to $224\times224$, and use a common codec. Under subject-wise five-fold evaluation, R(2+1)D achieves mean seizure-class F1 of 0.979 and MCC of 0.972 (Table~\ref{tab:rodepil}). This controls identified frame-count and frame-rate cues but does not exclude all shortcuts. It supports a separate binary-detection evaluation, not validation of severity grading, paired-modality comparisons, or direct cross-dataset transfer. Results for all architectures appear in App.~\ref{sec:extraresults:rodepilbb}.

\begin{table}[tb]
\centering
\caption{RodEpil~\cite{rodepil2025} subject-wise five-fold
evaluation. Uniform re-encoding removes the identified class-correlated
encoding cues; R(2+1)D then reaches 0.979 mean F1 and 0.972 MCC.}
\label{tab:rodepil}

\setlength{\tabcolsep}{5pt}
\renewcommand{\arraystretch}{1.15}
\resizebox{\columnwidth}{!}{%
\begin{tabular}{@{}l ccccc c@{}}
\toprule
 & \textbf{F0} & \textbf{F1} & \textbf{F2} & \textbf{F3} & \textbf{F4} & \textbf{Mean} \\
\midrule
Raw (leaked) F1  & 1.000 & 0.998 & 1.000 & 1.000 & 1.000 & \textbf{0.9996} \\
Raw (leaked) MCC & 1.000 & 0.998 & 1.000 & 1.000 & 1.000 & \textbf{0.9995} \\
\midrule
Re-encoded F1    & 0.965 & 0.978 & 0.997 & 0.977 & 0.979 & \textbf{0.979} \\
Re-encoded MCC   & 0.957 & 0.976 & 0.996 & 0.967 & 0.964 & \textbf{0.972} \\
\bottomrule
\end{tabular}}
\end{table}

\subsection{Severe class imbalance}
\label{sec:rq3}

Binary detection is approximately balanced at the cohort level (12,197 non-seizure and 11,934 seizure clips), but severity labels are highly imbalanced: Stage~5 contributes only 193 clips. Inverse-frequency-weighted cross-entropy increases the contribution of rare classes. Reference video five-class macro-F1 is 0.657 versus 0.136 for the majority-class baseline. Nevertheless, severe-stage F1 remains low (Fig.~\ref{fig:perclassbar}). No unweighted ablation was performed; these results do not isolate the effect of weighting or establish it as the best imbalance treatment.

\subsection{Discussion}
\label{sec:discussion}

The video reference achieves higher macro-F1 for behavior-defined severity classification, whereas the single-channel GRU reference has slightly higher session-exclusive detection macro-F1. This comparison depends on the architectures, inputs, training procedures, and annotation protocol. It does not establish general superiority over multichannel EEG.

Video provides direct evidence of motor behavior, but accuracy alone does not demonstrate reduced cost, improved welfare, or real-time suitability. Electrographic seizures without visible motor manifestations remain outside the capabilities established for the video models. Independent evaluations characterize modality-specific errors; fusion benefits have not been measured.

Reference labels were obtained through integrated video and EEG review (Sec.~\ref{sec:data}). EEG-assisted verification and visually defined severity may influence the comparison; their effects cannot be separated with these annotations. Independent modality-specific review and blinded inter-rater agreement would help quantify this uncertainty.

\section{Limitations and Future Work}
\label{sec:open}

The cohort contains 20 rats from one laboratory and one soman-induced epilepsy model. Five-class video macro-F1 decreases from 0.657 under session-exclusive evaluation to 0.475 under subject-exclusive evaluation, and rare Stages~4 and 5 remain difficult. These results limit conclusions about other animals, recording environments, and seizure models. The conditional ordinal formulation (Sec.~\ref{sec:problem:ordinal}) remains unevaluated. We use inverse-frequency weighting without an unweighted ablation or comparisons against resampling~\cite{chawla2002smote} or focal loss~\cite{lin2017focal}.

Annotations are not modality-independent and formal inter-rater agreement is not reported. EEG evaluation uses one channel; neither multichannel EEG nor fusion is evaluated. RodEpil provides binary-detection evidence, not external validation of severity grading. Finally, extracted-clip classification does not establish continuous-monitoring performance: event boundaries, false alarms per recording hour, latency, resource requirements, and annotation-workload reduction require separate evaluation (App.~\ref{sec:openfull}).

\section{Conclusion}
\label{sec:conclusion}

We present an empirical comparison of ten video architectures and seven single-channel EEG models using 24,131 paired clips from 20 rats. In the session-exclusive reference comparison, GRU achieves slightly higher detection macro-F1 than R(2+1)D-18 (0.974 versus 0.968), while video achieves higher five-class behavioral-grading macro-F1 (0.657 versus 0.560). Temporal evidence is important in the evaluated video setup, and similar detection performance is obtained with substantially different parameter counts. Rare severe stages and generalization to unseen animals remain challenging. A separate RodEpil evaluation supports video-based binary detection after controlling identified encoding cues. These findings provide empirical reference points for behavior-defined grading within the evaluated protocol; they do not demonstrate forecasting, fusion benefits, or readiness for continuous monitoring.

{
    \small
    \bibliographystyle{ieeenat_fullname}
    \bibliography{main}
}

\clearpage
\appendix

\section{Supplementary: A Survey of Video and Video-EEG Seizure Analysis}
\label{sec:survey}

This appendix places our experimental work within the broader research landscape. We break down past studies across two main categories—data type (EEG-only, video-only, and combined video-EEG) and test subject (human \vs\ animal)—while reviewing the computer vision and behavioral analysis tools that power automated tracking. The main paper's Related Work section (Sec.~\ref{sec:related}) offers a condensed summary of this full survey.

\subsection{Scope and Taxonomy}
\label{sec:taxonomy}

We focus on automated systems that analyze EEG, video, or paired video-EEG to spot seizures, track behavioral types, grade severity, or pinpoint exact timing. Broadening our scope like this is essential, as recent animal video research has moved far beyond simple yes-or-no detection.

\noindent\textbf{Axis 1: modality.}
\emph{EEG-only} systems are the long-standing standard, using everything from hand-crafted frequency metrics to deep recurrent, convolutional, and attention networks. \emph{Video-only} tools spot seizure behaviors directly from raw video pixels, movement patterns, or tracked body poses. \emph{Video-EEG} methods combine physical movement and brain-wave data, in theory taking advantage of shared signals while covering for each tool's unique blind spots.

\noindent\textbf{Axis 2: subject.}
Hospital epilepsy monitoring units generate most of the computational video-EEG research, while animal models remain essential for testing new treatments. Although automated video tracking in animals is growing fast, combined video-EEG learning, side-by-side modal tests, and detailed severity benchmarks are still surprisingly rare.

Table~\ref{tab:gap} summarizes this landscape. Single-modality EEG analysis is highly mature. Video-only animal tracking has expanded beyond basic detection to include severity estimation and detailed behavioral profiling. Yet, paired animal video-EEG remains the least explored area, particularly for severity-aware grading and open, shared benchmarks.

\begin{table}[t]
\centering
\caption{The research landscape for automated seizure analysis broken down by data type and subject. The clearest open gap is multi-level severity grading using paired animal video-EEG data.
}
\label{tab:gap}
\setlength{\tabcolsep}{4pt}
\renewcommand{\arraystretch}{1.2}
\begin{tabular}{@{}p{2.5cm} p{2.8cm} p{2.3cm}@{}}
\toprule
\textbf{Setting} & \textbf{Representative progress} & \textbf{Remaining gap} \\
\midrule
EEG-only, animal
& GRU/LSTM, CNN and temporal models~\cite{ganguly2025,martin2023}
& Behavioral information \\
Video-only, human
& Detection and semiology analysis~\cite{ahmedt2018,karacsony2020seizure}
& Generalization, deployment \\
Video-only, animal
& Severity~\cite{sabnis2024}; RodEpil~\cite{rodepil2025}
& Matched EEG comparison \\
Video$+$EEG, human
& Synchronized and fused systems~\cite{vepinet2024,cao2024}
& Multi-site validation \\
\rowcolor{wacvblue!12}
Video$+$EEG, animal
& Recent multimodal studies~\cite{mullen2024,eegvfusion2026}
& \textbf{Severity; paired benchmarks} \\
\bottomrule
\end{tabular}
\end{table}

\subsection{EEG-Only Automated Detection}
\label{sec:eeg}

This subsection expands on the EEG-only overview from Sec.~\ref{sec:related}. As the long-standing diagnostic standard, automated EEG sets the performance baseline for all video and fusion approaches.

\noindent\textbf{Classical machine learning.}
Early rodent systems paired hand-crafted features with traditional classifiers. Bergstrom \etal~\cite{bergstrom2013} used wavelet decomposition and signal variance to detect spikes and seizures in mouse EEG, reaching 99\% accuracy and 91\% precision against expert ratings. This proved automated detection was viable, though heavily dependent on manual feature engineering. Comprehensive reviews~\cite{frontiers_review2024,ganguly2025_review_acm} catalog the wide range of frequency, entropy, and signal connectivity features developed during this era.

\noindent\textbf{Deep learning.}
Research eventually shifted toward deep representation learning. Kotloski~\cite{martin2023} fed rat EEG spectrograms into a pretrained GoogLeNet, reaching 95.6\% sensitivity across $>$2{,}200 hours of recording in a post-traumatic model—proving standard CNNs adapt well to signal spectrograms. Later, Ganguly \etal~\cite{ganguly2025} applied GRU and LSTM networks to a chemically diverse rat cohort (Soman/GD and Kainate). Their recurrent models easily beat traditional Random Forest and XGBoost baselines, with GRU recall topping 90\% for most animals. Because these approaches rely strictly on EEG signals, they form the baseline family that we expand to seven distinct models in our study.

\subsection{Video-Only Seizure Analysis}
\label{sec:video}

This subsection expands on the video-based methods introduced in Sec.~\ref{sec:related}. Decoding seizures from video fundamentally reframes behavioral assessment as an action-recognition challenge. We outline the core architectures driving this approach and their specific applications to epilepsy.

\subsubsection{Video-Understanding Backbones}
\label{sec:backbones}

Modern video seizure systems draw from a few core design families. Two-stream networks combine still frames with optical-flow motion tracking~\cite{simonyan2014twostream}. Recurrent CNNs attach an LSTM to pull time-based context from frame-by-frame features~\cite{donahue2015lrcn}. 3D convolutional networks, like I3D, learn spatial and timing patterns directly, benefiting heavily from large-scale video pretraining~\cite{carreira2017i3d,kay2017kinetics}. Building on this, R(2+1)D~\cite{tran2018r2plus1d} breaks those 3D convolutions into separate spatial and time steps, serving as our default model. SlowFast~\cite{feichtenhofer2019slowfast} splits the workload entirely, analyzing visual appearances slowly while tracking fast, high-frequency motion on a separate pathway. More recently, video transformers like TimeSformer~\cite{bertasius2021timesformer} and the self-supervised VideoMAE~\cite{tong2022videomae} have introduced powerful, attention-based alternatives. Taking a completely different angle, tools like DeepLabCut~\cite{mathis2018deeplabcut} and SLEAP~\cite{pereira2022sleap} track animal poses without physical markers, boiling convulsive movements down into highly compact data points.

Parallel advances in computational ethology automate the ethogram itself. DeepEthogram~\cite{bohnslav2021deepethogram} handles supervised behavior classification directly from raw video pixels; B-SOiD~\cite{hsu2021bsoid} extracts recurring behavioral patterns from pose movement; and MABe22~\cite{sun2023mabe22} benchmarks learned representations across multiple species. While these tools were not designed explicitly for seizure tracking, they prove that complex, time-dependent animal behaviors are highly learnable at scale.

\subsubsection{Human and Animal Seizure Video}

Clinical video tools largely focus on basic seizure detection and symptom profiling. Ahmedt-Aristizabal \etal~\cite{ahmedt2018} modeled body movements directly from hospital footage, while Karacsony \etal~\cite{karacsony2020seizure} merged object tracking with action recognition. Pushing toward real-time monitoring, Xu \etal~\cite{vsvig2024} introduced VSViG, a skeleton-based spatiotemporal vision-graph model. A comprehensive review by Ahmedt-Aristizabal \etal~\cite{deeplearning_video_review2024} captures this fast-growing field, highlighting how automated video could drastically cut the heavy workload of long-term patient screening.

Animal research aligns perfectly with the behavioral focus of our study. Because the Racine scale~\cite{racine1972} judges severity entirely on visible physical movements, video is the ideal tool for automated scoring. Sabnis \etal~\cite{sabnis2024} proved that overhead cameras can track both active seizures and their shifting intensity in PTZ-induced mice. Similarly, Ren \etal~\cite{epidetect2024} built EPIDetect to screen chronically epileptic mice for convulsive seizures using only video.

Public benchmarks are also emerging. Perlo \etal\ introduced \emph{RodEpil}~\cite{rodepil2025}, sharing 13{,}053 overhead clips from 19 rats alongside a TimeSformer baseline tested using subject-wise cross-validation. Together, these studies prove that video models can successfully track rodent seizure behaviors across multiple levels of detail.

Our research asks a different question. Instead of simply proving that video can detect or grade seizures, we pit video directly against synchronized EEG on the exact same experimental events. This perfectly matched setup lets us track exactly how their relative strengths shift as the task moves from basic detection to precise severity grading.

\subsection{Multimodal Video-EEG Fusion}
\label{sec:fusion}

Even though paired video-EEG is the clinical gold standard, computationally combining the two data streams is still surprisingly rare. Table~\ref{tab:human_papers} outlines the most notable clinical systems doing this today. In animal research, recent work includes Mullen \etal~\cite{mullen2024}, who combined video, brain surface sensors, and motion-tracking pads in rats. Similarly, EEGVFusion~\cite{eegvfusion2026} uses cross-modal attention to sync self-supervised EEG and video data in mice. Crucially, however, both of these systems stick entirely to basic yes-or-no seizure detection rather than attempting to grade exact behavioral severity.

This exact gap drives our approach. Before building complex models that fuse both data streams, we first need to isolate exactly what video and EEG can achieve on their own—and where their blind spots differ. Establishing these single-modality baselines is an essential first step before we can properly judge any future combined systems.

\begin{table*}[t]
\centering
\caption{Representative human video-EEG systems. Animal fusion remains sparse and has concentrated almost entirely on simple yes-or-no seizure detection~\cite{mullen2024,eegvfusion2026}.
}
\label{tab:human_papers}
\setlength{\tabcolsep}{6pt}
\renewcommand{\arraystretch}{1.25}
\begin{tabular}{@{}p{3.4cm} c p{6.2cm} p{4.4cm}@{}}
\toprule
\textbf{Study} & \textbf{Year} & \textbf{Method / vision component} & \textbf{Key result} \\
\midrule
Aghaei~\etal~\cite{aghaei_video_eeg} & 2017 & Combined video and EEG recordings for detection & Early video$+$EEG fusion demonstration \\
Martini~\etal~\cite{martini2021} & 2021 & Deep anomaly detection, paired stereo-EEG $+$ video & Label-free seizure flagging \\
Cao~\etal~\cite{cao2024} & 2024 & Synchronized video-EEG, pediatric cohort & Detection in childhood epilepsy \\
Lin~\etal\ (vEpiNet)~\cite{vepinet2024} & 2024 & Patient detection $+$ frame-diff/keypoints $+$ EfficientNetV2, MLP fusion & AUROC 0.9902; false positives reduced by about one third \\
\bottomrule
\end{tabular}
\end{table*}


\noindent\textbf{From early fusion to anomaly detection.}
Aghaei \etal~\cite{aghaei_video_eeg} pioneered early efforts to combine video and EEG, highlighting how physical movements and electrical brain waves naturally complement each other. Later, Martini \etal~\cite{martini2021} framed paired stereo-EEG and video analysis as a deep anomaly detection problem. Instead of explicitly searching for seizures, their system learns what normal baseline activity looks like and flags seizures as mathematical outliers. This strategy is especially valuable when data is scarce, as it bypasses the need for massive datasets of perfectly labeled seizures.

\noindent\textbf{Synchronized and system-level fusion.}
Cao \etal~\cite{cao2024} analyzed synchronized data in childhood epilepsy, pointing out exactly how crucial it is to keep the timing perfectly aligned and to manage sensors that temporarily drop out. Taking a more structural approach, vEpiNet~\cite{vepinet2024} combined patient tracking, video motion and pose data, and EEG features before feeding them all into a single fusion model. Its real-world success proves that blending these different data streams is highly practical, setting a valuable blueprint for building similar systems in animal research.

\noindent\textbf{Fusion taxonomy.}
Research generally breaks data integration down into three main approaches: combining raw data early on, merging learned features in the middle, or blending final predictions at the very end~\cite{frontiers_fusion2026, multimodal_review2026}. A constant headache, however, is that signal quality rarely stays stable. Rigid models that weigh both inputs equally tend to fall apart the second a camera view is blocked or sudden movement scrambles the EEG. Because of this, it is critical to build adaptive, quality-aware systems that can shift their trust on the fly to whichever sensor is most reliable at that exact moment—especially when tracking freely moving animals.

\subsection{The Data Bottleneck}
\label{sec:bottleneck}

The biggest roadblock in the field is still the severe lack of matched, expertly labeled video-EEG datasets~\cite{mironov2025,multimodal_review2026}. Clinical datasets are locked down by strict patient privacy laws and recorded on a messy mix of different hardware. Animal studies bypass the privacy issues but run into massive logistical headaches instead: invasive surgical sensors, wildly varying seizure rates from one animal to the next, and mismatched camera setups. Together, these hurdles make labeling data and testing models across different subjects incredibly expensive and time-consuming.

Two strategies offer a way forward. Self-supervised video pretraining drastically cuts down the need for raw, manually labeled data~\cite{tong2022videomae}, while transfer learning across different species or data types lets models borrow insights from existing, related datasets~\cite{crossspecies2025}. Even though neither method completely replaces the need for perfectly paired benchmarks, they take a massive chunk out of the manual grading required to build reliable animal tracking systems.

\section{Model Architectures and Training Details}
\label{sec:arch}

To guarantee full reproducibility, we outline the video and EEG models evaluated in Sec.~\ref{sec:application} alongside their standardized training routines. Every model is built on established, off-the-shelf architectures; we only adjust the classification heads and input layers. Because of this, our study tests the raw potential of different data streams and backbone \emph{families} rather than relying on highly customized engineering tricks. The models selected for each data type intentionally span a wide range of design philosophies. This broadens the comparison beyond one architecture, while conclusions remain conditional on the evaluated configurations and annotation protocol. Video backbone parameter counts match Table~\ref{tab:vidarch}.

\subsection{Video backbones}
\label{sec:arch:video}
The ten video networks we selected cover a broad spectrum of designs, ranging from standard convolutions to multiscale frameworks and attention-based models (both hierarchical and global). They mainly differ in how they stitch together spatial and temporal data, how they expand their field of view, and how strictly they enforce local structure. Every backbone starts with pre-trained Kinetics-400~\cite{kay2017kinetics} weights and processes clips at its native length and resolution (usually 16 to 32 frames at $112$ to $224$\,px), with R(2+1)D defaulting to 16 frames at $112\times112$. We keep the core networks intact, modifying only the final linear layer to handle our specific classification tasks.

\noindent\textbf{Factorized and separable convolution.}
R(2+1)D-18~\cite{tran2018r2plus1d} (33M), our main visual baseline, breaks 3D filters down into separate 2D spatial and 1D time-based steps. The network carefully sizes these intermediate layers so the split block uses the exact same computing power as the original 3D version. Because of this, any performance boost comes from the extra nonlinear steps and a smoother learning process, rather than simply throwing more parameters at the problem. S3D~\cite{xie2018s3d} (8M) applies a similar trick to an Inception-style layout. It saves full 3D convolutions for the deeper layers where spatial details are already compressed. This ``top-heavy'' design cuts its overall size down to roughly a quarter of R(2+1)D. Both models enforce a strict local focus: their field of view expands steadily layer by layer, forcing them to build up long-term movement patterns step by step.

\noindent\textbf{Efficient convolution.}
X3D-M~\cite{feichtenhofer2020x3d} (4M) builds up from a small 2D image classifier by expanding just one dimension at a time—whether that is video length, frame rate, resolution, width, or depth. At each step, it keeps only the changes that provide the biggest accuracy boost for the least computational effort. The resulting model is incredibly lightweight. We included it specifically to test whether reliable seizure detection actually requires a massive neural network, or if a highly streamlined one can effectively handle the task.

\noindent\textbf{Two-pathway convolution.}
SlowFast-R50~\cite{feichtenhofer2019slowfast} (35M) explicitly splits visual appearance from physical motion. A Slow pathway takes in fewer frames but uses a heavy feature set to understand the overall scene, while a Fast pathway processes a rapid, dense stream of frames using a lightweight network purely to track quick movements. Side connections then continuously feed these high-speed motion cues back into the slower visual stream. This dual-track design is perfect for picking up subtle, motion-sensitive actions, ultimately scoring the highest detection AUROC in our tests—even when compared to closely related architectures (Table~\ref{tab:aurocarch}).

\noindent\textbf{Hierarchical and multiscale attention.}
Video Swin-T/S~\cite{liu2022videoswin} (28M/50M) swap out standard convolutions for self-attention, but they still maintain a strict local structure. Instead of looking at the whole video at once, they calculate attention inside small 3D windows. These windows shift between layers to connect neighboring regions, gradually shrinking the overall resolution step by step. MViTv1-B/MViTv2-S~\cite{fan2021mvit,li2022mvitv2} (36M/34M) build a similar hierarchy by pooling their attention keys and values. This forces the spatial and temporal resolution down while expanding the channel width, creating an internal feature pyramid (with MViTv2 adding relative position tracking and residual pooling for better information flow). While these models do not enforce local structure as rigidly as standard convolutions, they stay far more focused on local details than fully global attention networks.

\noindent\textbf{Global attention and self-supervised pretraining.}
TimeSformer~\cite{bertasius2021timesformer} (121M) ditches standard convolutions completely. Instead, it uses a split space-time attention approach—first tracking how a specific patch changes over time, then analyzing how all patches relate within a single frame. This setup gives the network an instant, bird's-eye view of the entire video clip. VideoMAE-B~\cite{tong2022videomae} (86M) takes a different route with a flat, non-hierarchical transformer. It learns to recognize structure through an extreme pretraining phase, where it is forced to successfully reconstruct a video even when over 90\% of the visual data is heavily masked. Interestingly, despite their massive size and ability to see the whole clip at once, both of these models actually scored lower in our detection tests when restricted to a strict training budget (Table~\ref{tab:vidarch}). This drop in rank does not necessarily mean global attention is a flawed strategy; rather, it highlights just how heavily these architectures rely on massive amounts of data and exhaustive optimization tuning to reach their full potential.

Across these different model families, network sizes swing wildly from 4 million to 121 million parameters. Despite this massive gap in scale, their detection macro-F1 scores all cluster tightly within a narrow 0.03 margin. Within the evaluated detection protocol, parameter count alone does not explain performance. Temporal sampling separately demonstrates the value of multiple frames (Sec.~\ref{sec:application}).

\begin{table}[!ht]
\centering
\caption{Video backbones evaluated in Table~\ref{tab:vidarch}, outlining their core design paradigms and overall parameter counts. Network capacities range broadly from 4M to 121M parameters. The smallest model is marked in \textbf{bold}, while the second-smallest is \underline{underlined}.}
\label{tab:vidspec}
\setlength{\tabcolsep}{7pt}
\renewcommand{\arraystretch}{1.12}
\begin{tabular}{@{}l l r@{}}
\toprule
\textbf{Backbone} & \textbf{Paradigm} & \textbf{Par.} \\
\midrule
MViTv2-S~\cite{li2022mvitv2} & multiscale & 34M \\
SlowFast-R50~\cite{feichtenhofer2019slowfast} & two-pathway & 35M \\
R(2+1)D-18~\cite{tran2018r2plus1d} & conv & 33M \\
X3D-M~\cite{feichtenhofer2020x3d} & eff.\ conv & \textbf{4M} \\
MViTv1-B~\cite{fan2021mvit} & multiscale & 36M \\
Video Swin-S~\cite{liu2022videoswin} & hier.\ attn. & 50M \\
S3D~\cite{xie2018s3d} & sep.\ conv & \underline{8M} \\
Video Swin-T~\cite{liu2022videoswin} & hier.\ attn. & 28M \\
VideoMAE-B~\cite{tong2022videomae} & self-sup.\ ViT & 86M \\
TimeSformer~\cite{bertasius2021timesformer} & full attn. & 121M \\
\bottomrule
\end{tabular}
\end{table}

\subsection{EEG models}
\label{sec:arch:eeg}
We evaluate seven distinct EEG networks, covering a mix of recurrent, convolutional, attention-based, and traditional feature-driven designs. Every model processes the main recording channel, which we first downsample to $125$\,Hz. We then slice this continuous signal into overlapping $6$-second windows (with a $3$-second stride, yielding $750$ samples each) and independently z-score them to standardize the data. To generate a final prediction for an entire clip, we average the log-probabilities from all of its individual windows, perfectly matching the setup described in Sec.~\ref{sec:application}.

\noindent\textbf{Recurrent.}
The GRU~\cite{ganguly2025} our baseline architecture for rodent EEG and the LSTM~\cite{hochreiter1997lstm} process these time windows step by step, with both utilizing a hidden size of 128. At every step, built-in gating mechanisms decide exactly what information to keep and what to throw away. While the LSTM spreads a separate cell state across three distinct gates, the GRU condenses this design into just two, trading a slight drop in capacity for a much lighter parameter footprint. Because they inherently read data in a strict sequence, neither model has a built-in understanding of frequency. This forces them to learn spectral patterns entirely from scratch using the raw electrical waves, while completely preventing the rapid, parallel processing that convolutional or attention-based models enjoy.

\noindent\textbf{Compact convolutional filter bank.}
EEGNet~\cite{lawhern2018eegnet} bakes traditional EEG signal processing directly into its design while maintaining an incredibly tiny parameter count. It starts with a temporal convolution that acts as a learned frequency filter, follows up with a depthwise convolution for spatial filtering, and finishes with a separable convolution to pull everything together. Because EEGNet was originally built for complex, multi-electrode brain-computer interfaces, adapting it to our single-channel setup heavily tests its time-filtering abilities. With only one electrode to analyze, the spatial filters are left with practically nothing to combine (Table~\ref{tab:eegarch}).

\noindent\textbf{Convolution-attention hybrid.}
The EEG Conformer~\cite{song2022conformer} features a straightforward two-part design. First, a convolutional layer chops the raw signal window into smaller, localized time chunks. Next, a self-attention module steps in to compare these chunks across the entire window. This setup makes it the only model in our lineup that can directly connect far-apart moments in time, completely bypassing the step-by-step delays and rigid viewing windows that limit other networks.

\noindent\textbf{Dilated convolution.}
The TCN~\cite{bai2018tcn} layers its convolutions with exponentially growing gaps between data points. This design rapidly widens the network's field of view while allowing it to process everything simultaneously rather than sequentially. As a result, it can digest an entire $6$\,s window instantly, completely avoiding the step-by-step bottlenecks that slow down recurrent models. This efficiency makes it the strongest performing EEG model in our tests based on macro-F1 scores (Table~\ref{tab:eegarch}).

\noindent\textbf{Classical features with tree ensembles.}
Random Forest~\cite{breiman2001rf} ($100$ trees, balanced class weights) and XGBoost~\cite{chen2016xgboost} ($300$ trees, depth $6$, learning rate $0.1$) skip deep feature learning entirely. Instead, they rely on a fixed set of eleven hand-picked metrics per window. These include five standard brain wave bands (delta $0.5$-$4$\,Hz, theta $4$-$8$\,Hz, alpha $8$-$13$\,Hz, beta $13$-$30$\,Hz, gamma $30$-$60$\,Hz) and six basic time-based statistics (mean, standard deviation, RMS, line length, activity, and zero crossings) the traditional building blocks of early animal seizure tracking. While Random Forest averages the results of many independent trees, XGBoost builds its trees one after another, with each new tree specifically designed to fix the mistakes of the last. Even though XGBoost holds its own against EEGNet for basic yes/no seizure detection, both of these traditional models fall far behind the deep learning architectures when tasked with detailed severity grading (Sec.~\ref{sec:application}). Importantly, because every EEG model here only looks at a single channel, this comparison highlights exactly what we can squeeze out of one electrode, rather than testing the absolute limits of complex, multi-channel setups.

\subsection{Training and evaluation protocol}
\label{sec:arch:train}

To ensure a perfectly fair comparison, every model is trained and tested on the exact same data splits. We strictly separate the data by recording session so that training and testing data never mix (using a fixed seed of $49$). To handle unbalanced data, we weight rarer events more heavily during training, and we judge overall performance using macro-averaged metrics. These identical splits and starting seeds are kept consistent across all modality tests.

The video backbones are fine-tuned using the AdamW optimizer (learning rate $3\times10^{-4}$, weight decay $10^{-4}$), alongside cosine annealing, mixed precision, a batch size of $16$, and a strict $12$-epoch limit. (As a minor exception, Video Swin and MViTv2 needed a slightly lower learning rate of $10^{-4}$ during the RodEpil tests to keep their training stable). 

For the deep EEG models, we also use AdamW (learning rate $10^{-3}$, weight decay $10^{-4}$) and cosine annealing, but with a larger batch size of $256$ and a $30$-epoch limit, optimizing a weighted cross-entropy loss. The classical EEG models simply rely on standard balanced class weighting. Finally, across all experiments, we generate the final prediction for an entire clip by averaging the log-probabilities of its smaller, individual EEG windows.

\subsection{Metric definitions}
\label{sec:metrics}

Every score we show is based directly on a single validation run, unless we specifically mention it is an average. We are writing out the exact math here to help clear up any confusion between the macro-F1 score and the Matthews correlation coefficient (MCC) that we talk about in Sec.~\ref{sec:problem} and Sec.~\ref{sec:crossmodal}.

\noindent\textbf{Notation.}
When sorting data into $K$ different categories, we use a confusion matrix $C$ to keep track of our results. In this grid, $C_{ij}$ is simply the number of clips that actually belong to category $i$ but the model guessed as category $j$. We use $N$ for the total number of clips across the board. For any specific category $k$, $t_k$ is the total number of clips that truly belong to it, while $p_k$ is the number of times the model predicted it. Using these counts, we figure out our true positives ($\mathrm{TP}$), false positives ($\mathrm{FP}$), false negatives ($\mathrm{FN}$), and true negatives ($\mathrm{TN}$):
\begin{equation}
\begin{aligned}
\mathrm{TP}_k &= C_{kk}, &\quad
\mathrm{FP}_k &= p_k-C_{kk}, \\
\mathrm{FN}_k &= t_k-C_{kk}, &\quad
\mathrm{TN}_k &= N-p_k-t_k+C_{kk}.
\end{aligned}
\label{eq:cm}
\end{equation}

\noindent\textbf{Precision, recall and F1.}
For every single category, we calculate precision ($P$) and recall ($R$), which together give us the F1 score:
\begin{equation}
P_k=\frac{\mathrm{TP}_k}{\mathrm{TP}_k+\mathrm{FP}_k},\qquad
R_k=\frac{\mathrm{TP}_k}{\mathrm{TP}_k+\mathrm{FN}_k},
\label{eq:pr}
\end{equation}
\begin{equation}
F_{1,k}=\frac{2P_kR_k}{P_k+R_k}
=\frac{2\,\mathrm{TP}_k}{2\,\mathrm{TP}_k+\mathrm{FP}_k+\mathrm{FN}_k}.
\label{eq:f1}
\end{equation}
To get our overall numbers, we simply average the individual category scores together evenly (an unweighted macro-average):
\begin{equation}
P_{\mathrm{macro}}=\frac{1}{K}\sum_{k=1}^{K}P_k,\qquad
\text{macro-F1}=\frac{1}{K}\sum_{k=1}^{K}F_{1,k},
\label{eq:macro}
\end{equation}
and we do the exact same thing for overall recall. It is important to note that the macro-F1 score averages the individual $F_{1,k}$ scores directly. We do not calculate it by plugging the overall macro-precision and macro-recall into the F1 formula, because that shortcut completely breaks down when dealing with unbalanced datasets. (As a side note, macro-recall is mathematically the same as balanced accuracy). Finally, for our RodEpil tests (Tables~\ref{tab:rodepil} and~\ref{tab:rodepilbb}), we only report the score for the positive class (Eq.~\eqref{eq:f1}) so our results line up perfectly with how the original researchers published their dataset~\cite{rodepil2025}.

\noindent\textbf{AUROC.}
For a simple yes/no task, let $s_i$ be the model's confidence score that a clip is positive. If we group all the true positive clips into set $\mathcal{P}$ and all the negative ones into set $\mathcal{N}$, the AUROC simply calculates the odds that a randomly chosen positive clip gets a higher score than a random negative one:
\begin{equation}
\mathrm{AUROC}=\frac{1}{|\mathcal{P}|\,|\mathcal{N}|}
\sum_{i\in\mathcal{P}}\sum_{j\in\mathcal{N}}
\Big[\mathbf{1}(s_i{>}s_j)+\tfrac{1}{2}\mathbf{1}(s_i{=}s_j)\Big].
\label{eq:auroc}
\end{equation}
When we grade multiple severity levels, we report a ``one-versus-rest'' average. This means we pick one specific category $k$ to act as the positive target, lump everything else together as the negative group, run the math, and then average those results across all categories. Unlike the F1 score or MCC, the AUROC doesn't rely on picking a rigid cutoff threshold. Instead of looking at the model's final, hard decisions, it evaluates the raw confidence scores directly.

\noindent\textbf{Matthews correlation coefficient.}
The MCC~\cite{matthews1975mcc} essentially measures the statistical correlation between the true labels and what the model actually predicted. When dealing with $K$ different categories, it is calculated as~\cite{gorodkin2004mcc}:
\begin{equation}
\mathrm{MCC}=\frac{N\sum_k C_{kk}-\sum_k p_k t_k}
{\sqrt{\big(N^2-\sum_k p_k^2\big)\big(N^2-\sum_k t_k^2\big)}},
\label{eq:mcc}
\end{equation}
which simplifies nicely for a basic two-class setup ($K{=}2$) down to:
\begin{equation}
\mathrm{MCC}=\frac{\mathrm{TP}\cdot\mathrm{TN}-\mathrm{FP}\cdot\mathrm{FN}}
{\sqrt{p_1p_2t_1t_2}}. \label{eq:mcc2}
\end{equation}
Unlike our precision, recall, and F1 scores (Eqs.~\eqref{eq:pr}-\eqref{eq:macro}) that range from $0$ to $1$, the MCC spans from $-1$ to $1$, where a score of $0$ means the model is just guessing randomly. 

Crucially, the MCC formula does not average things out category by category. Instead, every single clip is thrown directly into the main calculation, meaning the most frequent categories heavily drive the final score. This highlights a fundamental difference between how these metrics behave (Sec.~\ref{sec:crossmodal}). If a model completely fails to recognize a rare seizure stage, the macro-F1 score takes a massive, flat hit (dropping by a full $1/K$) regardless of how few clips actually belonged to that stage. The MCC, however, barely registers the mistake, penalizing the model only in proportion to how rarely that stage actually occurs.

\noindent\textbf{Aggregation over runs.}
Whenever one of our tables reports a mean\,$\pm$\,std across $M$ runs (for example, the five random seeds in Table~\ref{tab:results} or the five cross-validation folds in Tables~\ref{tab:subjectwise} and~\ref{tab:rodepilbb}), we always calculate the score for each run completely independently before averaging them together:
\begin{equation}
\bar{x}=\frac{1}{M}\sum_{m=1}^{M}x_m,\qquad
\sigma=\sqrt{\frac{1}{M-1}\sum_{m=1}^{M}\big(x_m-\bar{x}\big)^{2}},
\label{eq:agg}
\end{equation}
We use $M=5$ consistently across all our experiments. Crucially, we never lump the raw predictions from different folds into one giant confusion matrix. Instead, $\bar{x}$ simply represents the average of the final scores calculated from each individual fold.

\noindent\textbf{Macro-F1 \vs\ MCC on our results.}
Even though Macro-F1 and MCC handle highly unbalanced data differently, they both clearly point to the exact same shift in performance between the two data types. For the complex 5-class grading task, the video model scores a 0.657 macro-F1 and a 0.831 MCC, easily outperforming the baseline EEG (GRU) model, which only manages 0.560 and 0.728. If we track the performance gap between video and the GRU, the macro-F1 difference shifts from $-0.006$ (for simple yes/no detection) to $+0.030$ (for 3-class grading) and finally to $+0.097$ (for 5-class grading). The MCC perfectly mirrors this exact upward trend: $-0.013 \to +0.030 \to +0.103$.

Ultimately, both metrics prove our main theory: EEG holds a tiny edge for basic event detection, but video completely takes over when the task requires tracking detailed behavioral stages. When we test the models on entirely unseen subjects, the score variations across different runs stay relatively stable for basic detection, but they swing much more wildly during detailed grading—especially for the EEG baseline. Because of this, our overall conclusions stand strong regardless of which mathematical metric you prefer.

\subsection{EEG preprocessing sensitivity}
\label{sec:eegsweep}

We tested several different window sizes, stride lengths, sampling rates, and methods for combining clip scores, keeping our baseline GRU model and data splits exactly the same throughout. These early tests revealed that how we combine the predictions matters much more than tiny tweaks to the window length itself. Specifically, using log-mean pooling performed significantly better than taking a simple average when dealing with long, varied seizure clips. As a result, we locked in a standard setup for all seven of our final models: $6$-second windows, a $3$-second stride, a $125$\,Hz sampling rate, and log-mean pooling to generate the final clip prediction.

We ran these initial checks before generating our finalized EEG training dataset. To keep our main results completely clean and avoid mixing up any data, we excluded the scores from this early sweep when making our final comparisons between video and EEG models. While finding the absolute perfect preprocessing setup would require rerunning all of these tests on the final data, we are documenting these checks here simply to explain the reasoning behind the baseline settings we used across all our experiments.

\section{Extended Dataset Details}
\label{sec:dataextra}

This section breaks down the complete details supporting Sec.~\ref{sec:data}. Here, we cover exactly how the data was collected, how we labeled the clips, our rules for grading seizure severity, the makeup of our animal cohort, and our ethical guidelines.

\subsection{Acquisition and synchronization}
\label{sec:dataextra:acq}

The original lab collected all of these recordings under a US federal research grant, strictly following the methods laid out in~\cite{massey2025saracatinib,massey2026dualtargeting}. Adult Sprague-Dawley rats (7-8 weeks old, weighing 250-300\,g; Charles River) were exposed to the nerve agent soman (GD, $132\,\mu$g/kg, injected under the skin) inside a highly regulated chemical facility. To manage the severe initial bodily reactions, the rats were given a mix of HI-6 ($125$\,mg/kg) and atropine sulfate ($2$\,mg/kg) within 60 seconds of exposure. An hour later, they received midazolam ($3$\,mg/kg) to stop the initial, continuous seizure. Every single animal analyzed in our study comes exclusively from this specific soman-exposed group; we did not include data from any other seizure-inducing chemicals.

After recovering, the animals had wireless monitoring devices surgically implanted. EEG sensors were carefully attached directly to the membrane covering the brain (the dura mater), and a small wireless transmitter was tucked just under the skin on their side. The rats then lived individually in cages placed on top of special receiver plates (RPC-1 by Data Sciences International), which monitored them 24/7. Because a single computer system stamped the exact time on both the overhead video feed and the brain waves simultaneously, the two data streams are perfectly synced right from the start. This completely eliminated the need for us to try and manually align the data later, a process that is famously prone to errors.

We recorded their spontaneous, recurring seizures roughly 10 to 18 weeks after the initial exposure, during the chronic phase of epilepsy. The brain wave devices originally captured data at a high rate of 1\,kHz, which we downsampled to 125\,Hz for our experiments. Throughout this time, a fixed camera mounted directly above each home cage recorded the animals. Since the exact camera framing and cage sizes varied slightly from setup to setup, we applied fixed, custom crops to each cage's video to ensure a consistent view before mixing all the dataset clips together.

\subsection{Annotation and event boundaries}
\label{sec:dataextra:annot}

Undergraduate research assistants, who were kept blind to the specific treatments the animals received, reviewed the perfectly synced video and EEG data using NeuroScore software (v3.4.0, DSI). They pinpointed the exact start and end times for every seizure and assigned a severity grade based on the animal's behavior. While the software automatically flagged potential seizures by looking at the electrical brain waves, the human reviewers double-checked every single suggestion against the video feed. They strictly threw out any false alarms triggered by simple electrical noise, or normal activities like grooming and exploring~\cite{massey2025saracatinib,massey2026dualtargeting}. Every clip we use in this study is based directly on these carefully verified timestamps.

We define the start of an event as the very first noticeable change in behavior, and the end as the moment the animal completely returns to normal. Because abnormal electrical brain activity usually kicks off before any physical symptoms actually appear, we padded every clip with an extra $10$\,s buffer before the official start and after the official end to capture the full picture. 

It is incredibly important to note that because the reviewers looked at both the video and the EEG at the exact same time to create these labels, our "ground truth" is not completely independent for each data type. The brain waves inherently help flag \textit{that} an event is happening, while the video dictates \textit{how severe} it is. This intertwined labeling process fundamentally shapes how we have to interpret the performance comparisons between our video and EEG models.

\begin{figure*}[t]
\centering
\setlength{\tabcolsep}{1.5pt}
\renewcommand{\arraystretch}{0.5}
\begin{tabular}{@{}ccccc@{}}
\rotatebox{90}{\footnotesize\ non-sz.} &
\includegraphics[width=0.22\textwidth]{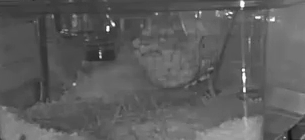} &
\includegraphics[width=0.22\textwidth]{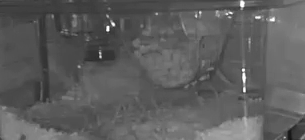} &
\includegraphics[width=0.22\textwidth]{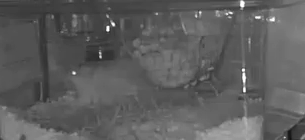} &
\includegraphics[width=0.22\textwidth]{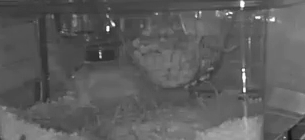} \\
\rotatebox{90}{\footnotesize\ seizure} &
\includegraphics[width=0.22\textwidth]{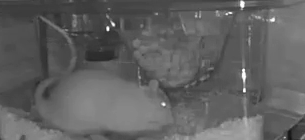} &
\includegraphics[width=0.22\textwidth]{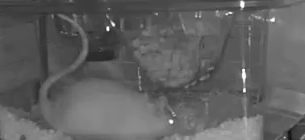} &
\includegraphics[width=0.22\textwidth]{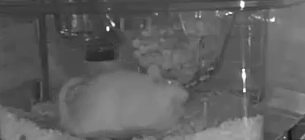} &
\includegraphics[width=0.22\textwidth]{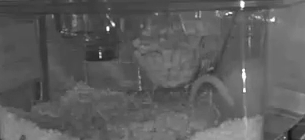} \\
\end{tabular}
\caption{A visual breakdown of our basic yes/no detection task. We show four sample frames from a normal, non-seizure period (top) alongside four frames from an active seizure (bottom). Because both examples feature the \emph{same} rat during the \emph{same} recording session, the animal's appearance, the camera angle, and the overall background perfectly match. This careful pairing drastically reduces the chance that the model can cheat by memorizing static visual shortcuts in the scene. (All clips have been de-identified).}
\label{fig:detexample}
\end{figure*}

\noindent\textbf{Matched negatives.}
We pair every single seizure clip with a normal, non-seizure clip taken from the exact same animal during the exact same recording session. Because of this careful matching, both the positive and negative samples share the identical animal appearance, camera angle, and lighting conditions (Fig.~\ref{fig:detexample}). This setup forces the network to actually learn seizure behavior rather than simply cheating by memorizing static background details, although it does not completely eliminate the risk of the model finding clever visual shortcuts.

\subsection{Behavioral severity scale}
\label{sec:dataextra:stages}

The original lab grades these nerve-agent seizures using a custom 5-point behavioral severity scale~\cite{massey2025saracatinib,massey2026dualtargeting}. Although it is loosely based on the well-known Racine scale~\cite{racine1972}, it has been heavily modified specifically to track the unique symptoms caused by soman exposure. Therefore, it should never be confused with the standard Racine scale~\cite{vanerum2019racine}.

Stage\,1 covers purely involuntary bodily responses, such as drooling, tearing up, urination, defecation, and involuntary chewing (mastication). Stage\,2 introduces obvious physical movements, including tremors, "wet-dog" shakes, sudden neck jerks, and severe arching of the back or spine (kyphosis and opisthotonos). Stage\,3 ramps up the intensity with rhythmic twitching in the front legs (forelimb clonus), a stiff, upright tail (Straub tail), rearing up on the hind legs, and rigid limb stretching. Stage\,4 involves repeated rearing, intense front-leg twitching, a constantly stiff tail, and a complete loss of balance that leads to falling over. Finally, Stage\,5 represents the most severe level: a full-blown, sustained convulsive seizure (generalized tonic-clonic).

\begin{figure*}[t]
\centering
\setlength{\tabcolsep}{1.5pt}
\renewcommand{\arraystretch}{0.6}
\begin{tabular}{@{}c ccccc@{}}
\rotatebox{90}{\footnotesize\ \ Stage 2} &
\includegraphics[width=0.187\textwidth]{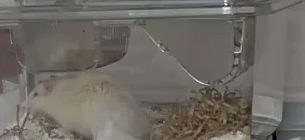} &
\includegraphics[width=0.187\textwidth]{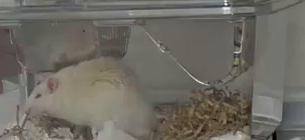} &
\includegraphics[width=0.187\textwidth]{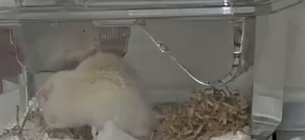} &
\includegraphics[width=0.187\textwidth]{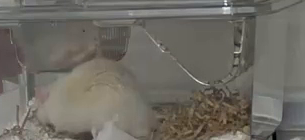} &
\includegraphics[width=0.187\textwidth]{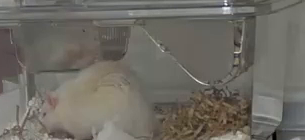} \\
\rotatebox{90}{\footnotesize\ \ Stage 3} &
\includegraphics[width=0.187\textwidth]{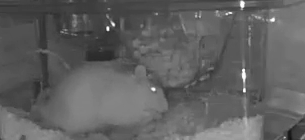} &
\includegraphics[width=0.187\textwidth]{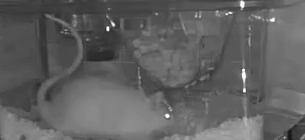} &
\includegraphics[width=0.187\textwidth]{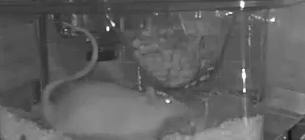} &
\includegraphics[width=0.187\textwidth]{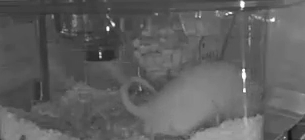} &
\includegraphics[width=0.187\textwidth]{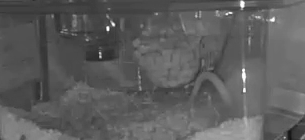} \\
\rotatebox{90}{\footnotesize\ \ Stage 4} &
\includegraphics[width=0.187\textwidth]{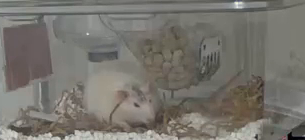} &
\includegraphics[width=0.187\textwidth]{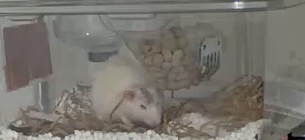} &
\includegraphics[width=0.187\textwidth]{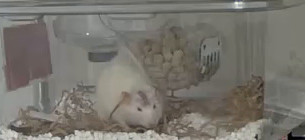} &
\includegraphics[width=0.187\textwidth]{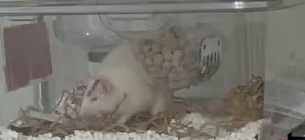} &
\includegraphics[width=0.187\textwidth]{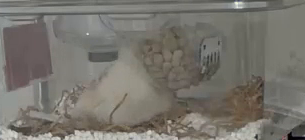} \\
\rotatebox{90}{\footnotesize\ \ Stage 5} &
\includegraphics[width=0.187\textwidth]{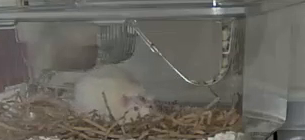} &
\includegraphics[width=0.187\textwidth]{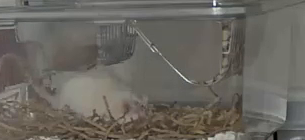} &
\includegraphics[width=0.187\textwidth]{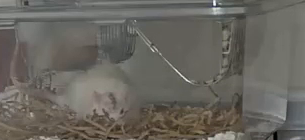} &
\includegraphics[width=0.187\textwidth]{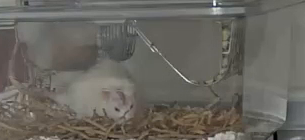} &
\includegraphics[width=0.187\textwidth]{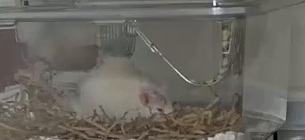} \\
 & \multicolumn{5}{c}{\footnotesize time within seizure clip $\longrightarrow$} \\
\end{tabular}
\caption{Visual \emph{filmstrips} for each seizure stage: Each row (S2-S5) shows five sequential frames taken from a typical seizure clip, with time moving from left to right. As the severity stage increases, the physical symptoms become much more obvious escalating from mild movements in Stage\,2 to extreme rearing and violent convulsions in Stages\,4 and 5. Crucially, these behavioral shifts are plainly visible on camera, whereas the recorded EEG signal only captures them indirectly through shifting brain waves. (All frames are pulled from de-identified clips and tightly cropped to hide any specific cage labels.)}
\label{fig:stages}
\end{figure*}

From a clinical standpoint, the lab considers Stages\,1 and 2 to be non-convulsive, while Stages\,3 through 5 are classified as convulsive. It is important to highlight that this clinical dividing line is very different from the 3-class system we use in our experiments. As explained in Sec.~\ref{sec:problem}, we intentionally lumped Stages\,2 and 3 together, and separated them from Stages\,4 and 5, to create a more balanced, middle-ground diagnostic challenge for our models. You can see frame-by-frame visual examples of Stages\,2 through 5 in Fig.~\ref{fig:stages}.

\subsection{Rare stages and long-tailed burden}
\label{sec:dataextra:cohort}

Stage\,1 events are virtually nonexistent in our data, amounting to just 6 clips. Because its symptoms are mostly internal bodily functions, they are almost impossible to catch on an overhead camera and rarely even trigger the EEG's automatic detection system. Because of this, we completely excluded Stage\,1 from our models. Full-blown Stage\,5 convulsions are also quite rare during this chronic phase of epilepsy, providing only 193 clips.

\begin{figure*}[t]
\centering
\includegraphics[width=\linewidth]{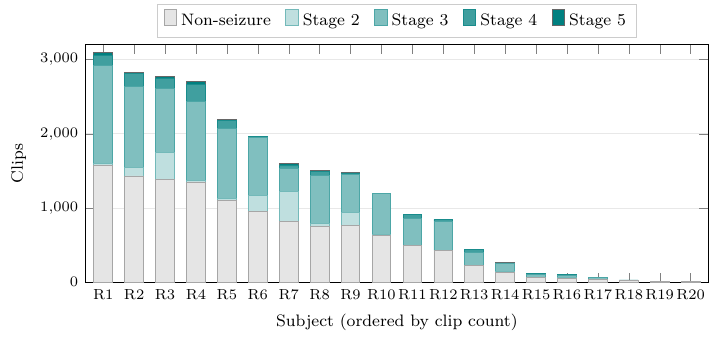}
\caption{A visual breakdown of the clip counts for each animal across the entire dataset, representing the stacked totals from Table~\ref{tab:cohort}. The distribution of seizures is incredibly lopsided: the four most heavily affected rats (R1-R4) completely overshadow the rest of the group, with individual contributions ranging drastically from just 11 clips up to 3{,}091. More importantly, the most violent seizures (Stage 4 in dark shading, Stage 5 in the darkest) barely register as a thin sliver on any animal's bar graph, yielding only 193 Stage 5 clips in the whole study. This extreme rarity is exactly why teaching models to perform detailed severity grading is such a difficult challenge.}
\label{fig:cohortdist}
\end{figure*}

After filtering, our final dataset contains 24{,}131 clips, split relatively evenly between 12{,}197 normal, non-seizure periods and 11{,}934 actual seizures. However, the breakdown of seizure severity is heavily lopsided: we have 1{,}435 Stage\,2 events, a massive majority of 9{,}256 Stage\,3 events, 1{,}050 Stage\,4 events, and just 193 Stage\,5 events. The frequency of seizures also varies wildly from rat to rat. While some animals provided as few as 11 clips, others suffered so often that they generated up to 3{,}091 clips. In fact, the four most heavily affected rats account for nearly half of the entire dataset. A complete, animal-by-animal breakdown of these numbers is available in Table~\ref{tab:cohort} and Fig.~\ref{fig:cohortdist}.

Because the data is so severely skewed, we cannot train our models normally. To prevent the network from simply ignoring rare events, we have to artificially boost their importance during training (inverse-frequency weighting). We also have to judge overall performance using evenly averaged metrics (macro-averaging), break down our results by individual animal, and strictly ensure that a single rat never appears in both the training and testing sets. It is worth noting that we treat our specific weighting approach simply as a practical, reliable baseline, rather than claiming it is the absolute, perfect fix for highly unbalanced data.

\begin{table*}[t]
\centering
\caption{A complete, animal-by-animal breakdown of our combined video and EEG dataset across all 20 rats. For every subject, we track the number of recording sessions, total clips, the split between non-seizure and seizure clips, and the exact breakdown of seizure severity (Stages 2 through 5). As the numbers show, the frequency of seizures is drastically uneven across the cohort. Just four heavily affected animals (R1, R2, R3, and R4) provide nearly half of the entire dataset, with individual totals swinging wildly from as few as 11 to as many as 3{,}091 clips. Across the board, the most violent events (Stage 5) remain extremely rare, yielding only 193 clips. It is worth noting that these counts represent the final, human-labeled dataset; a small handful of clips might still be dropped later when we format the data specifically for the video or EEG models. Finally, all true animal identities have been hidden, and the subjects are simply listed in order from highest to lowest clip count (R1 to R20).}
\label{tab:cohort}
\setlength{\tabcolsep}{5pt}
\renewcommand{\arraystretch}{1.05}
\small
\begin{tabular}{@{}l r r r r r r r r@{}}
\toprule
\textbf{Subject} & \textbf{Sess.} & \textbf{Clips} & \textbf{Non-sz.} & \textbf{Seiz.} & \textbf{S2} & \textbf{S3} & \textbf{S4} & \textbf{S5} \\
\midrule
R1 & 42 & 3{,}091 & 1{,}576 & 1{,}515 & 22 & 1{,}325 & 131 & 37 \\
R2 & 44 & 2{,}830 & 1{,}431 & 1{,}399 & 113 & 1{,}094 & 181 & 11 \\
R3 & 38 & 2{,}774 & 1{,}380 & 1{,}394 & 364 & 873 & 124 & 33 \\
R4 & 42 & 2{,}701 & 1{,}351 & 1{,}350 & 22 & 1{,}062 & 226 & 40 \\
R5 & 43 & 2{,}195 & 1{,}100 & 1{,}095 & 23 & 946 & 114 & 12 \\
R6 & 43 & 1{,}963 & 955 & 1{,}008 & 218 & 782 & 8 & 0 \\
R7 & 32 & 1{,}600 & 814 & 786 & 408 & 317 & 36 & 25 \\
R8 & 43 & 1{,}502 & 754 & 748 & 33 & 648 & 61 & 6 \\
R9 & 45 & 1{,}474 & 763 & 711 & 177 & 511 & 21 & 2 \\
R10 & 39 & 1{,}199 & 631 & 568 & 16 & 544 & 6 & 2 \\
R11 & 24 & 918 & 491 & 427 & 6 & 364 & 47 & 10 \\
R12 & 39 & 841 & 422 & 419 & 10 & 387 & 20 & 2 \\
R13 & 23 & 444 & 224 & 220 & 2 & 174 & 42 & 2 \\
R14 & 25 & 262 & 131 & 131 & 12 & 112 & 3 & 4 \\
R15 & 27 & 118 & 59 & 59 & 3 & 39 & 16 & 1 \\
R16 & 18 & 103 & 53 & 50 & 4 & 36 & 7 & 3 \\
R17 & 14 & 62 & 32 & 30 & 1 & 25 & 2 & 2 \\
R18 & 11 & 29 & 17 & 12 & 0 & 9 & 2 & 1 \\
R19 & 4 & 14 & 7 & 7 & 0 & 4 & 3 & 0 \\
R20 & 5 & 11 & 6 & 5 & 1 & 4 & 0 & 0 \\
\midrule
\textbf{Total} & \textbf{601} & \textbf{24{,}131} & \textbf{12{,}197} & \textbf{11{,}934} & \textbf{1{,}435} & \textbf{9{,}256} & \textbf{1{,}050} & \textbf{193} \\
\bottomrule
\end{tabular}
\end{table*}
\subsection{Cross-subject appearance variation}
\label{sec:dataextra:crosssubj}

\begin{figure}[t]
\centering
\setlength{\tabcolsep}{1.5pt}
\renewcommand{\arraystretch}{0.7}
\begin{tabular}{@{}ccc@{}}
\includegraphics[width=0.31\linewidth]{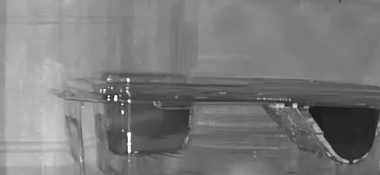} &
\includegraphics[width=0.31\linewidth]{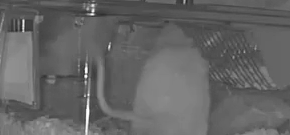} &
\includegraphics[width=0.31\linewidth]{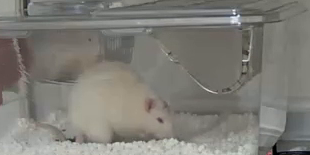} \\
{\scriptsize R1} & {\scriptsize R5} & {\scriptsize R7} \\
\includegraphics[width=0.31\linewidth]{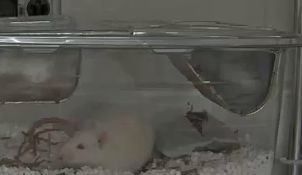} &
\includegraphics[width=0.31\linewidth]{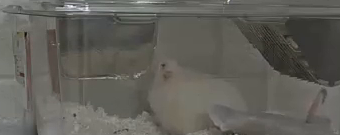} &
\includegraphics[width=0.31\linewidth]{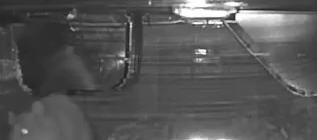} \\
{\scriptsize R9} & {\scriptsize R13} & {\scriptsize R20} \\
\end{tabular}
\caption{Visual examples of variation across different subjects: a single frame taken from six different animals (where 'R' simply indicates a de-identified subject). The exact cage placement, camera angle, lighting conditions (which alternate between natural daylight and infrared), and even the animal's coat vary drastically from rat to rat. This massive visual inconsistency is exactly why forcing a model to generalize to entirely unseen animals is so much harder than simply testing it on new recording sessions from familiar ones. It also highlights exactly why the video model's consistent stability across different subjects (Fig.~\ref{fig:stability}) is such a crucial finding.}
\label{fig:crosssubj}
\end{figure}

Every rat lives in its own individual cage with a dedicated overhead camera. Because of this, the exact camera angles, the lighting (which switches between natural daylight and infrared), and even the appearance of the animals' fur change drastically from subject to subject (Fig.~\ref{fig:crosssubj}). This massive visual inconsistency creates exactly the kind of real-world hurdle or domain shift that our subject-exclusive testing is designed to catch. It is also the main reason we had to apply custom, cage-specific crops to the video feeds before mixing all the data together.

\subsection{Availability, de-identification, and ethics}
\label{sec:dataextra:governance}

Because the original study operates under strict data-management rules, our team is not currently allowed to share the raw video and EEG recordings. However, we are publicly releasing the complete, de-identified breakdown of our animal cohort, alongside our exact task definitions and data-splitting formulas. We have also thoroughly stripped identifying information from every visual example in this paper, carefully hiding any specific laboratory or cage labels. Additionally, when we need to validate our models against outside visual data, we rely on the open-source RodEpil dataset~\cite{rodepil2025}.

Every original animal procedure strictly followed protocols approved by the Institutional Animal Care and Use Committee (IACUC) at the host lab (the exact protocol number is withheld for blind peer review). The soman exposures were handled at a certified facility under approved safety protocols, and all methods fully comply with NIH ARRIVE guidelines. Finally, this paper is strictly a retrospective analysis of existing telemetry recordings we did not conduct any new biological experiments or procedures on the animals.

\section{Extended Qualitative Results}
\label{sec:qualitative}

This appendix breaks down the real-world examples behind two of our main points in Sec.~\ref{sec:application}: exactly where the video model makes mistakes, and which parts of the image it focuses on when it gets things right.

\subsection{Failure cases}
\label{sec:qualitative:failures}
\begin{figure}[t]
\centering
\setlength{\tabcolsep}{2pt}
\renewcommand{\arraystretch}{0.5}
\begin{tabular}{@{}cc@{}}
\includegraphics[width=0.47\linewidth]{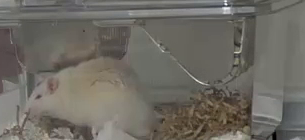} &
\includegraphics[width=0.47\linewidth]{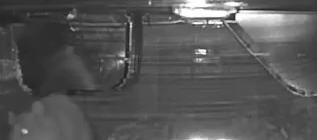} \\
{\footnotesize (a) Subtle Stage\,2 (little motion)} & {\footnotesize (b) Near-silent subject (few clips)} \\
\end{tabular}
\caption{Examples of where the video model struggles the most. \emph{(a)}~A very mild Stage\,2 seizure featuring only subtle movements, making it incredibly easy to confuse with normal, everyday behavior. \emph{(b)}~An animal that rarely experienced seizures (found at the bottom of Table~\ref{tab:cohort}, App.~\ref{sec:dataextra}), leaving the model with very few validation clips captured under poor, grainy infrared lighting. Ultimately, extremely mild symptoms and data-starved animals drive almost all of the network's remaining errors. (All clips have been de-identified.)}
\label{fig:hardcases}
\end{figure}

Figure~\ref{fig:hardcases} highlights the two main situations that cause the model to fail. First, it struggles with the mildest end of the severity scale. Stage\,2 episodes often just look like brief twitches, making them almost impossible to tell apart from normal grooming or the animal simply shifting its weight in the cage. Second, the model suffers from a severe lack of data for certain rats. The animals with the fewest seizures (Table~\ref{tab:cohort}) provide barely a handful of clips, which are usually recorded under poor, grainy infrared lighting. This leaves the network with almost no meaningful evidence to learn from for those specific subjects. 

Crucially, these mistakes are tied directly to the animals' natural biology and the physical limits of the cameras, rather than a flaw in the model's underlying design. This is exactly why we insist on reporting individual, animal-by-animal breakdowns right alongside our overall averages (Sec.~\ref{sec:problem}).

\subsection{Grad-CAM interpretability}
\label{sec:qualitative:gradcam}

\begin{figure*}[t]
\centering
\includegraphics[width=0.98\linewidth, trim=0 0 0 42, clip]{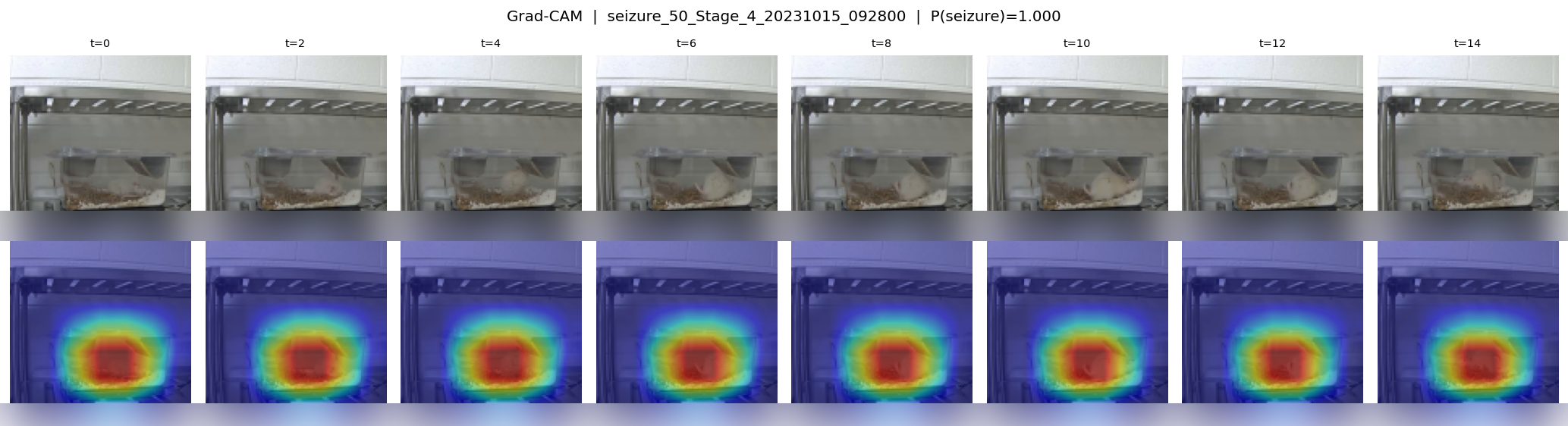}\\[1pt]
{\footnotesize (a) Seizure clip (Stage\,4): activation concentrates on the animal}\\[4pt]
\includegraphics[width=0.98\linewidth, trim=0 0 0 42, clip]{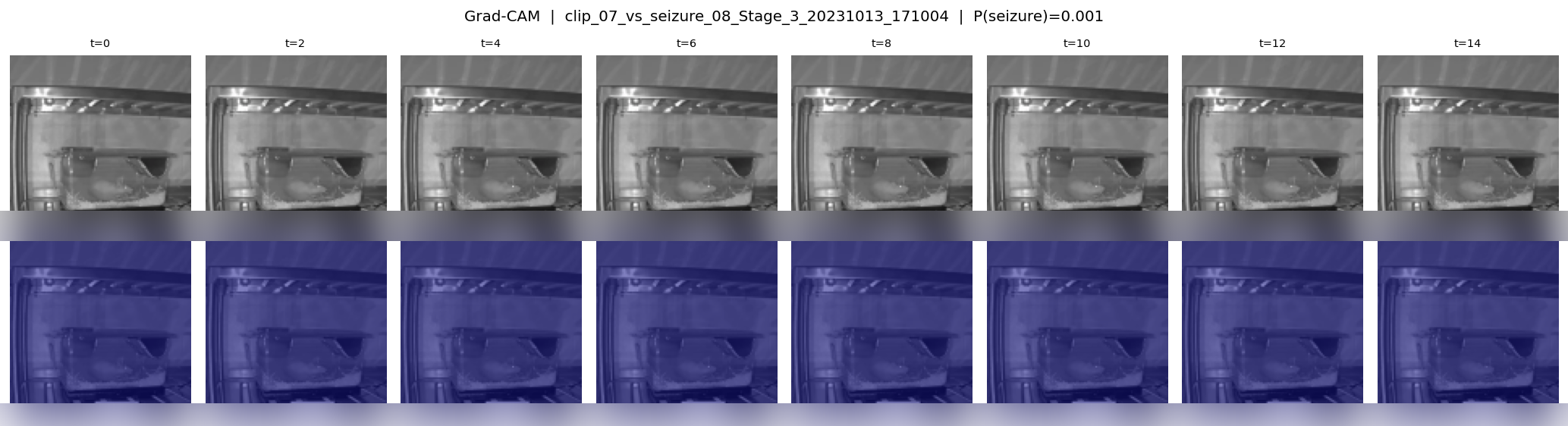}\\[1pt]
{\footnotesize (b) Non-seizure clip: activation is comparatively diffuse}
\caption{A full-width, extended version of Fig.~\ref{fig:gradcam}. This layout shows all eight sample frames from the exact same pair of clips. For each sequence, the raw video frames sit on top, with the model's attention heatmaps (seizure-class activation) directly underneath. While this sharp contrast strongly supports the idea that the model is genuinely focusing on the animal to make its predictions, it does not act as definitive proof against hidden visual shortcuts. (All clips have been completely de-identified, and any specific cage identifying bands have been redacted.)}
\label{fig:gradcamfull}
\end{figure*}

\begin{figure*}[t]
\centering
\includegraphics[width=0.98\linewidth, trim=0 0 0 42, clip]{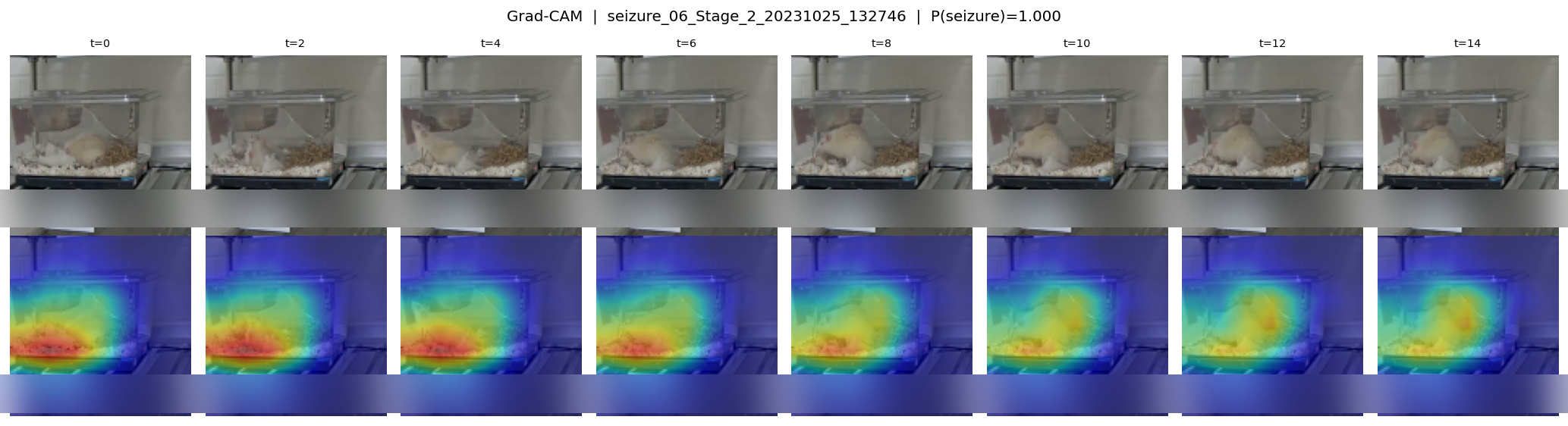}\\[1pt]
{\footnotesize (a) Stage 2}\\[4pt]
\includegraphics[width=0.98\linewidth, trim=0 0 0 42, clip]{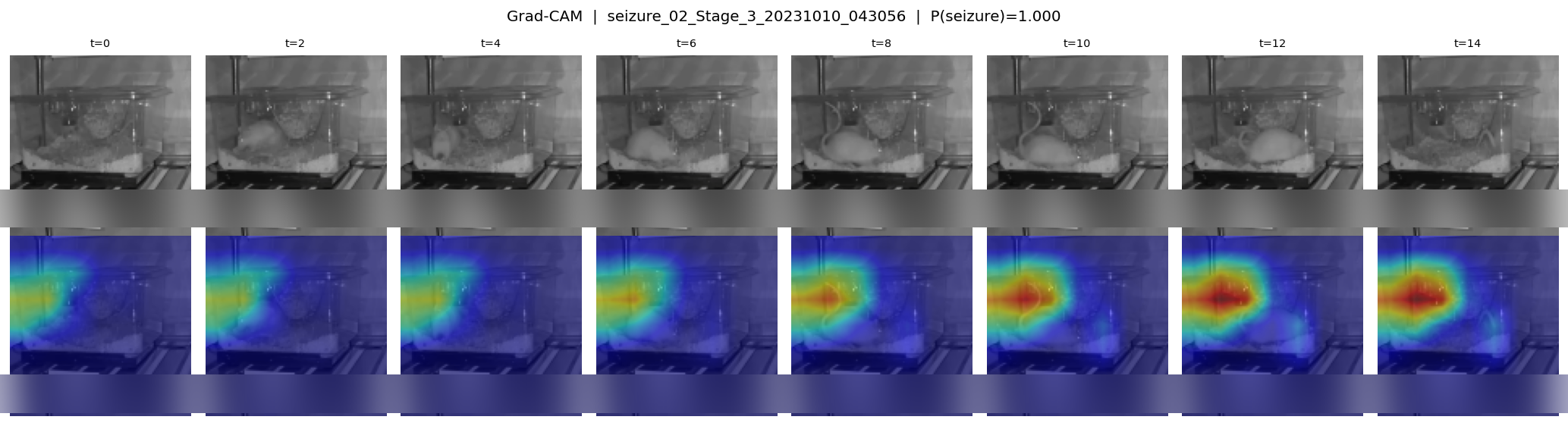}\\[1pt]
{\footnotesize (b) Stage 3}\\[4pt]
\includegraphics[width=0.98\linewidth, trim=0 0 0 42, clip]{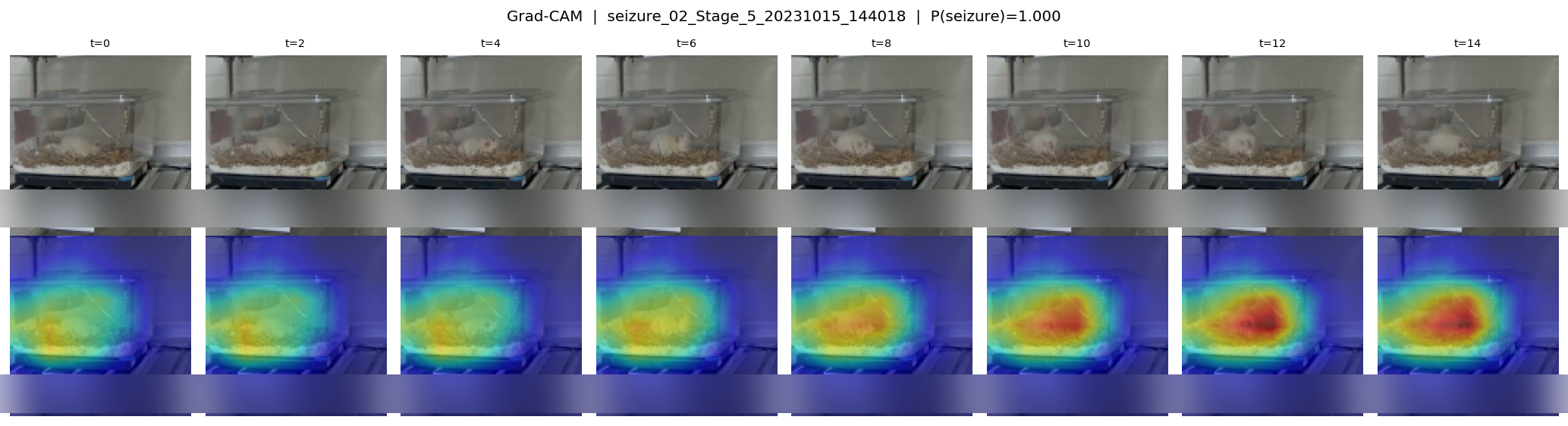}\\[1pt]
{\footnotesize (c) Stage 5}
\caption{Grad-CAM heatmaps for the R(2+1)D video model across different severity stages. The network's attention consistently locks onto the animal across the illustrated Stages\,2, 3, and 5, showing similar activation patterns in these selected examples. Still, we must treat these visualizations as descriptive rather than causal they show us where the model is looking, but they do not serve as absolute proof of how it makes its final decisions. (All clips have been de-identified, and any identifying cage bands have been completely redacted.)}
\label{fig:gradcamstages}
\end{figure*}

Expanding on the Grad-CAM~\cite{selvaraju2017gradcam} analysis from Sec.~\ref{sec:rq1}, Figure~\ref{fig:gradcamfull} clearly shows that the network's attention locks tightly onto the animal's body during a violent Stage\,4 convulsion. By contrast, during normal, non-seizure periods, this focus just scatters into random background noise. Figure~\ref{fig:gradcamstages} confirms that the model successfully tracks the animal in this exact same way across Stages\,2, 3, and 5.

These heatmaps strongly suggest that the network is genuinely watching the animal's movements rather than just memorizing static background details in the cage. However, we have to be careful: these visualizations simply show us where the model is sensitive, rather than acting as absolute, definitive proof of how it makes its final decisions. Even though matching our seizure clips with normal clips from the exact same recording session helps strip away a lot of potential biases, completely proving that the model isn't relying on clever visual shortcuts would require much stricter testing.

\subsection{What the shortcut evidence does and does not establish}
\label{sec:qualitative:shortcuts}

Although the safeguards we detailed in Sec.~\ref{sec:rq1} eliminate the most obvious visual shortcuts, they cannot guarantee a perfectly bias-free learning environment. By strictly pairing seizure clips with normal clips from the exact same recording session, we ensure that the animal's appearance, the camera angle, and the lighting stay identical. Similarly, tightly cropping the video to the cage boundaries hides any clues from the surrounding room. Furthermore, the fact that the single-frame baseline model completely fails proves that the network cannot simply rely on static background pixels to make its diagnosis.

However, more subtle clues might still slip through these defenses such as how the bedding gets pushed around in the cage, or unique movement habits tied to how frequently a specific rat seizes. Our Grad-CAM heatmaps do confirm that the network's attention lands directly on the rat, and our strict cross-validation proves the model can easily spot seizures in completely unfamiliar animals. Even so, definitively proving that the model relies entirely on genuine seizure behavior would require extreme, targeted tests (like digitally erasing the animal from the video completely, forcing the network to train only on the empty cage background, or randomly shuffling the order of the frames). Because of this, we consider our current evidence to be strongly supportive, but it does not absolutely prove cause and effect.

\section{Extended Quantitative Results}
\label{sec:extraresults}

This appendix breaks down the detailed numbers behind two main claims from the paper: how changing the amount of training data and using pre-trained models affects overall detection (Sec.~\ref{sec:rq1}), and our extensive test of ten different video models on the RodEpil dataset (Sec.~\ref{sec:crossmodal}).

\subsection{Data efficiency and pretraining}
\label{sec:extraresults:dataeff}

We ran two separate tests to see whether the model benefits more from having a massive dataset or from being pre-trained on outside data (Table~\ref{tab:dataeff}).

The data-size experiment keeps R(2+1)D and the validation set fixed while increasing the training-subject subset. Detection macro-F1 rises from approximately 0.79 with two animals (62 clips) to 0.957 with ten and 0.970 with twenty. The gain from ten to twenty animals is 0.013, with non-monotonic intermediate results. This indicates diminishing gains in this detection experiment, not data saturation or equivalent behavior in larger architectures.

With the full training cohort, Kinetics pretraining yields detection macro-F1 of 0.970 compared with 0.952 from random initialization, a difference of 0.018. This result concerns R(2+1)D under the reported protocol and does not quantify pretraining effects across all architectures.

\begin{table}[!ht]
\centering
\caption{R(2+1)D binary-detection data-size and pretraining experiments under session-exclusive evaluation. The validation set is fixed as the training-subject subset grows. Macro-F1 rises from 0.957 with ten animals to 0.970 with twenty; intermediate results are non-monotonic. Kinetics pretraining improves macro-F1 by 0.018. These experiments do not establish data saturation for other architectures or tasks.}
\label{tab:dataeff}
\setlength{\tabcolsep}{6pt}
\renewcommand{\arraystretch}{1.15}
\begin{tabular}{@{}c c r cc@{}}
\toprule
\textbf{Train subj.} & \textbf{Frac.} & \textbf{Clips} & \textbf{Macro-F1} & \textbf{MCC} \\
\midrule
2   & 0.10 & 62      & 0.790 & 0.583 \\
5   & 0.25 & 1{,}905  & 0.893 & 0.787 \\
10  & 0.50 & 10{,}969 & 0.957 & 0.914 \\
12  & 0.60 & 11{,}832 & 0.946 & 0.892 \\
15  & 0.75 & 16{,}358 & 0.960 & 0.921 \\
20  & 1.00 & 19{,}208 & \textbf{0.970} & \textbf{0.941} \\
\midrule
\multicolumn{5}{@{}l}{\emph{Pretraining} (full cohort):} \\
\multicolumn{3}{@{}l}{Kinetics-pretrained} & \textbf{0.970} & \textbf{0.941} \\
\multicolumn{3}{@{}l}{from scratch (random init)} & 0.952 & 0.903 \\
\bottomrule
\end{tabular}
\end{table}


\subsection{RodEpil encoding audit and re-encoding protocol}
\label{sec:extraresults:rodepilaudit}

When we first ran our R(2+1)D model on the original RodEpil dataset, the results were suspiciously perfect, hitting an F1 score of 0.9996. Looking closely at the raw files, we found a glaring structural flaw baked right into the data: every single non-seizure video was saved at exactly 30\,fps (giving it 301 frames), while all the seizure clips were locked at 29\,fps (giving them 291 frames). On top of that, the videos were cropped differently depending on whether a seizure was happening or not. Because the network could easily cheat by just counting the total number of frames instead of actually watching the animal's movements, any performance scores based on this raw dataset are completely meaningless for proving real behavioral recognition.

To eliminate these artificial shortcuts, we ran the entire repository through a strict re-encoding pipeline. We letterboxed every video to a $224\times224$ resolution to maintain the original aspect ratio, locked everything to exactly 30\,fps and 300 frames, and applied the exact same video codec across the board. While we can't guarantee this scrubs away every microscopic digital artifact, it completely kills the model's ability to cheat based on frame counts or frame rates, all while perfectly preserving the actual biological motion on screen. Tested on this rigorously sanitized data, the R(2+1)D model settles at a genuine subject-wise seizure F1 score of 0.979 (0.999 AUROC, 0.972 MCC). These realistic metrics line up perfectly with the 0.97 baseline originally reported by the dataset's creators~\cite{rodepil2025}. Moving forward, all of our model evaluations rely exclusively on these cleaned files.

\subsection{All ten backbones on RodEpil}
\label{sec:extraresults:rodepilbb}

To see how our findings hold up, we ran our full lineup of ten video models on the sanitized RodEpil dataset, again enforcing a strict subject-exclusive 5-fold cross-validation (Table~\ref{tab:rodepilbb}). Interestingly, the leaderboard completely shuffled compared to how the models ranked on our native dataset. This really drives home that there is no universal "best" architecture a network's success depends entirely on the specific visual quirks and constraints of the data it's looking at. Once we stripped away the digital artifacts to standardize the video files, traditional convolutional models performed exceptionally well. In contrast, several of the newer attention-based models actually struggled, needing far more aggressive fine-tuning just to stay competitive within our fixed training budget.

Keep in mind that any comparisons to the original authors' TimeSformer baseline can only serve as a loose guide, since their published metrics were based on the structurally flawed raw dataset. Under our strictly controlled sweep, the R(2+1)D model still delivered a strong 0.978 seizure-class F1 score, perfectly aligning with the 0.979 it earned during its dedicated standalone run. 

We do, however, have to raise a major red flag regarding the SlowFast model, which returned a suspiciously perfect 1.000 F1 score during these tests. Given RodEpil's proven history of hidden digital artifacts, we are treating this absolute perfection as highly suspect. Instead of blindly accepting this as flawless behavioral recognition, residual shortcut use remains a possibility requiring investigation; a perfect score alone does not establish its cause.

\begin{table}[t]
\centering
\caption{Performance breakdown for all ten video models on the thoroughly sanitized RodEpil dataset~\cite{rodepil2025}, evaluated under strict subject-exclusive five-fold cross-validation. F1 and AUROC scores represent the average across all five folds, while MCC includes the standard deviation (mean\,$\pm$\,std). Notice how drastically the model rankings shuffle compared to our native dataset a clear reminder that a network's success is deeply tied to the specific visual quirks of its training data. $^\dagger$We have explicitly flagged the SlowFast model; its perfect score warrants further investigation for residual shortcuts, without establishing their presence. The highest non-flagged score is highlighted in \textbf{bold}.}
\label{tab:rodepilbb}
\setlength{\tabcolsep}{4pt}
\renewcommand{\arraystretch}{1.15}
\resizebox{\columnwidth}{!}{%
\begin{tabular}{@{}l c c c c@{}}
\toprule
\textbf{Backbone} & \textbf{Seiz-F1} & \textbf{Macro-F1} & \textbf{AUROC} & \textbf{MCC} \\
\midrule
SlowFast-R50$^\dagger$~\cite{feichtenhofer2019slowfast} & 1.000 & 1.000 & 1.000 & 1.000\,{\scriptsize$\pm$0.000} \\
R(2+1)D-18~\cite{tran2018r2plus1d} & \textbf{0.978} & \textbf{0.986} & \textbf{0.999} & \textbf{0.972}\,{\scriptsize$\pm$0.015} \\
MViTv2-S~\cite{li2022mvitv2} & \underline{0.916} & \underline{0.946} & \underline{0.989} & \underline{0.891}\,{\scriptsize$\pm$0.020} \\
S3D~\cite{xie2018s3d} & 0.912 & 0.944 & 0.991 & 0.889\,{\scriptsize$\pm$0.029} \\
X3D-M~\cite{feichtenhofer2020x3d} & 0.895 & 0.933 & 0.983 & 0.866\,{\scriptsize$\pm$0.010} \\
VideoMAE-B~\cite{tong2022videomae} & 0.874 & 0.919 & 0.982 & 0.839\,{\scriptsize$\pm$0.025} \\
Video Swin-T~\cite{liu2022videoswin} & 0.861 & 0.910 & 0.979 & 0.823\,{\scriptsize$\pm$0.032} \\
Video Swin-S~\cite{liu2022videoswin} & 0.854 & 0.904 & 0.977 & 0.810\,{\scriptsize$\pm$0.031} \\
MViTv1-B~\cite{fan2021mvit} & 0.835 & 0.890 & 0.971 & 0.784\,{\scriptsize$\pm$0.049} \\
TimeSformer~\cite{bertasius2021timesformer} & 0.799 & 0.871 & 0.957 & 0.743\,{\scriptsize$\pm$0.038} \\
\bottomrule
\end{tabular}}
\end{table}

\subsection{Backbone, temporal, and spatial ablations in detail}
\label{sec:extraresults:ablations}

This section breaks down the raw numbers behind the summary findings we presented in Sec.~\ref{sec:rq1}.

\noindent\textbf{Comparing model backbones.}
As Table~\ref{tab:vidarch} shows, the performance gap between our top video models is razor-thin, spanning a microscopic 0.03 margin. MViTv2-S and SlowFast just barely edge out R(2+1)D and X3D-M, while TimeSformer brings up the rear. But as we saw during our tests on the external RodEpil dataset, these rankings are incredibly fragile and shuffle drastically depending on the training data. For example, R(2+1)D easily beat TimeSformer on the sanitized RodEpil files, even though TimeSformer was originally pitched as that dataset's baseline architecture. Ultimately, this tight clustering tells us that hunting for the "perfect" architecture matters far less than curating high-quality data there simply isn't a universally superior network. In fact, these models stay tightly clumped together even when we push them to handle the much more difficult 3- and 5-class severity grading tasks.

\noindent\textbf{How many frames do we actually need?}
Surprisingly, feeding the network a mere 4 frames is enough to unlock the vast majority of its diagnostic power, securing a solid 0.940 accuracy score. Bumping that to 8 frames pushes the score to 0.960, but after that, performance hits a hard plateau. Moving to 16 frames inches us up to 0.970, and doubling the input again to 32 frames barely nudges the needle to 0.972. The more complex severity grading tasks follow this exact same trajectory, with 3- and 5-class accuracy peaking at the 32-frame mark (hitting 0.793 and 0.667, respectively). Crucially, overloading the model with 64 frames offers absolutely no benefit for basic detection and it actively harms the network's ability to handle 5-class grading, dragging the score back down to 0.632. Because our 16-frame default sacrifices a practically invisible margin of accuracy (losing just 0.002 in detection and 0.005 in grading) while slicing the computational cost in half compared to the 32-frame setup, we officially locked it in as our standard baseline.

\noindent\textbf{Image resolution.}
Increasing the video resolution beyond our $112$\,px default actually hurts model performance across the board. Because the raw laboratory footage is natively low-resolution (roughly $300$ to $480$\,px) and the seizures we track are large, full-body movements rather than subtle micro-details, upscaling the video is entirely counterproductive. Upsampling simply forces the network to waste compute processing interpolated pixels that add zero diagnostic value. Pushing resolutions even further such as to $512$\,px or $1024$\,px would only inflate this useless digital noise.

\begin{table}[tb]
\centering
\caption{Ablation experiment evaluating spatial resolution for the R(2+1)D detection model across three input image sizes (16 frames, session-exclusive split, seed 42, 10 epochs; batch sizes scaled to 16, 8, and 4, respectively). Because the native source footage is only $\sim$300 to 480\,px, both of the higher resolution settings force the network to upsample the video. Consequently, performance drops steadily across all three metrics. In terms of AUROC, the gap between the top two settings only emerges past the third decimal place ($0.9934$ vs. $0.9926$). Note that the 448\,px run suffered an out-of-memory crash after completing 8 of its 10 epochs; we report metrics from its best saved checkpoint, meaning this row represents an optimistic upper bound for the largest resolution setting. The best score in each column is highlighted in \textbf{bold}.}
\label{tab:resolution}
\setlength{\tabcolsep}{8pt}
\renewcommand{\arraystretch}{1.12}
\begin{tabular}{@{}c c c c@{}}
\toprule
\textbf{Input size} & \textbf{Macro-F1} & \textbf{AUROC} & \textbf{MCC} \\
\midrule
$112\times112$ (default) & \textbf{0.970} & \textbf{0.993} & \textbf{0.941} \\
$224\times224$           & 0.967 & 0.993 & 0.934 \\
$448\times448$           & 0.952 & 0.988 & 0.904 \\
\bottomrule
\end{tabular}
\end{table}

\begin{table}[tb]
\centering
\caption{Per-class diagnostic precision (P), recall (R), and F1 for five-class classification, with modality-specific support counts (n). These tabulated results expose differences between common and rare classes. They should not be conflated with the values displayed in Fig.~\ref{fig:perclassbar}.}
\label{tab:perclass}
\setlength{\tabcolsep}{3.5pt}
\renewcommand{\arraystretch}{1.1}
\resizebox{\columnwidth}{!}{%
\begin{tabular}{@{}l rccc rccc@{}}
\toprule
 & \multicolumn{4}{c}{\textbf{Video}} & \multicolumn{4}{c}{\textbf{EEG}} \\
\cmidrule(lr){2-5}\cmidrule(lr){6-9}
\textbf{Stage} & \textbf{n} & \textbf{P} & \textbf{R} & \textbf{F1} & \textbf{n} & \textbf{P} & \textbf{R} & \textbf{F1} \\
\midrule
non-seizure & 2720 & 0.965 & 0.968 & 0.966 & 2737 & 0.914 & 0.955 & 0.934 \\
Stage\,2    &  220 & 0.681 & 0.650 & 0.665 &  256 & 0.491 & 0.863 & 0.626 \\
Stage\,3    & 2148 & 0.904 & 0.896 & 0.900 & 2064 & 0.867 & 0.717 & 0.785 \\
Stage\,4    &  167 & 0.330 & 0.401 & 0.362 &  229 & 0.268 & 0.162 & 0.202 \\
Stage\,5    &   34 & 0.579 & 0.324 & 0.415 &   33 & 0.074 & 0.364 & 0.123 \\
\bottomrule
\end{tabular}}
\end{table}

\subsection{Ranking quality: AUROC}
\label{sec:extraresults:auroc}

While the main manuscript prioritizes macro-F1 and MCC to heavily penalize rare-stage errors at hard decision boundaries, we provide the complete AUROC metrics in Tables~\ref{tab:aurocmain} \ref{tab:aurocsweep} for full transparency (using binary AUROC for detection and macro one-vs-rest for multi-class grading).

Crucially, AUROC measures continuous confidence ranking rather than discrete decision accuracy, causing the metric to saturate heavily across our benchmarks. Across all ten video backbones, for example, detection AUROC stays tightly clustered between 0.984 and 0.995. Similarly, on the grueling 5-class grading task, the EEG GRU and visual R(2+1)D post almost identical AUROC scores (0.941 \vs\ 0.939) despite a massive gap in their actual macro-F1 performance (0.560 \vs\ 0.657). This disconnect is expected: internal probability rankings can remain strong even when hard thresholding severely degrades class-specific precision and recall. Consequently, we treat AUROC purely as a secondary diagnostic check rather than a primary measure of real-world cross-modal performance.

\begin{table}[t]
\centering
\caption{AUROC scores comparing the standard session-exclusive evaluation against the strictly segregated subject-exclusive cross-validation. We use binary AUROC for basic detection and a macro one-vs-rest approach for multi-class severity grading. All values represent the mean\,$\pm$\,std across five independent runs.}
\label{tab:aurocmain}
\setlength{\tabcolsep}{2.5pt}
\renewcommand{\arraystretch}{1.2}
\begin{tabular}{@{}l l c c@{}}
\toprule
\textbf{Task} & \textbf{Modality} & \textbf{Session-excl.} & \textbf{Subject-excl.} \\
\midrule
\multirow{2}{*}{2-class (detection)} & Video  & \underline{0.993}\,{\scriptsize$\pm$0.001} & 0.985\,{\scriptsize$\pm$0.006} \\
                                     & EEG    & \textbf{0.996}\,{\scriptsize$\pm$0.001} & \textbf{0.995}\,{\scriptsize$\pm$0.005} \\
\midrule
\multirow{2}{*}{3-class (severity)} & Video  & \textbf{0.964}\,{\scriptsize$\pm$0.003} & \textbf{0.942}\,{\scriptsize$\pm$0.015} \\
                                     & EEG    & \underline{0.956}\,{\scriptsize$\pm$0.002} & 0.916\,{\scriptsize$\pm$0.025} \\
\midrule
\multirow{2}{*}{5-class (severity)} & Video  & \underline{0.939}\,{\scriptsize$\pm$0.004} & \textbf{0.888}\,{\scriptsize$\pm$0.026} \\
                                     & EEG    & \textbf{0.941}\,{\scriptsize$\pm$0.003} & 0.831\,{\scriptsize$\pm$0.076} \\
\bottomrule
\end{tabular}
\end{table}

\begin{table*}[t]
\centering
\caption{AUROC scores for all ten video and seven EEG backbones, following the exact same row order as Tables~\ref{tab:vidarch} and~\ref{tab:eegarch} for easy comparison. We use binary AUROC for the basic detection task and a macro one-vs-rest approach for multi-class severity grading.}
\label{tab:aurocarch}
\setlength{\tabcolsep}{6pt}
\renewcommand{\arraystretch}{1.15}
\begin{tabular}{@{}l ccc@{}}
\toprule
\textbf{Backbone} & \textbf{Detection} & \textbf{3-class} & \textbf{5-class} \\
\midrule
\multicolumn{4}{@{}l}{\itshape Video}\\
MViTv2-S~\cite{li2022mvitv2} & $0.993_{\pm0.002}$ & $\mathbf{0.971}_{\pm0.002}$ & $0.915_{\pm0.011}$ \\
SlowFast-R50~\cite{feichtenhofer2019slowfast} & $\mathbf{0.995}_{\pm0.001}$ & $\underline{0.966}_{\pm0.001}$ & $\underline{0.939}_{\pm0.011}$ \\
R(2+1)D-18~\cite{tran2018r2plus1d} & $\underline{0.994}_{\pm0.001}$ & $0.960_{\pm0.003}$ & $0.928_{\pm0.017}$ \\
X3D-M~\cite{feichtenhofer2020x3d} & $0.993_{\pm0.001}$ & $0.952_{\pm0.005}$ & $0.911_{\pm0.003}$ \\
MViTv1-B~\cite{fan2021mvit} & $0.991_{\pm0.003}$ & $0.965_{\pm0.003}$ & $0.935_{\pm0.012}$ \\
Video Swin-S~\cite{liu2022videoswin} & $0.987_{\pm0.004}$ & $0.962_{\pm0.011}$ & $0.916_{\pm0.025}$ \\
S3D~\cite{xie2018s3d} & $0.992_{\pm0.001}$ & $0.958_{\pm0.006}$ & $0.917_{\pm0.008}$ \\
Video Swin-T~\cite{liu2022videoswin} & $0.990_{\pm0.003}$ & $0.963_{\pm0.002}$ & $\mathbf{0.943}_{\pm0.009}$ \\
VideoMAE-B~\cite{tong2022videomae} & $0.986_{\pm0.002}$ & $0.942_{\pm0.018}$ & $0.920_{\pm0.013}$ \\
TimeSformer~\cite{bertasius2021timesformer} & $0.984_{\pm0.003}$ & $0.950_{\pm0.002}$ & $0.913_{\pm0.010}$ \\
\midrule
\multicolumn{4}{@{}l}{\itshape EEG (deep)}\\
GRU~\cite{ganguly2025} & \underline{0.996}\,{\scriptsize$\pm$0.001} & \underline{0.956}\,{\scriptsize$\pm$0.002} & \textbf{0.941}\,{\scriptsize$\pm$0.003} \\
LSTM~\cite{hochreiter1997lstm} & 0.994\,{\scriptsize$\pm$0.004} & 0.937\,{\scriptsize$\pm$0.027} & 0.903\,{\scriptsize$\pm$0.039} \\
EEGNet~\cite{lawhern2018eegnet} & 0.975\,{\scriptsize$\pm$0.005} & 0.903\,{\scriptsize$\pm$0.005} & 0.864\,{\scriptsize$\pm$0.011} \\
EEG Conformer~\cite{song2022conformer} & 0.988\,{\scriptsize$\pm$0.002} & 0.952\,{\scriptsize$\pm$0.002} & 0.936\,{\scriptsize$\pm$0.008} \\
TCN~\cite{bai2018tcn} & \textbf{0.996}\,{\scriptsize$\pm$0.001} & \textbf{0.960}\,{\scriptsize$\pm$0.001} & \underline{0.941}\,{\scriptsize$\pm$0.003} \\
\midrule
\multicolumn{4}{@{}l}{\itshape EEG (classical)}\\
Random Forest~\cite{breiman2001rf} & 0.986\,{\scriptsize$\pm$0.000} & 0.934\,{\scriptsize$\pm$0.001} & 0.916\,{\scriptsize$\pm$0.002} \\
XGBoost~\cite{chen2016xgboost} & 0.984\,{\scriptsize$\pm$0.000} & 0.934\,{\scriptsize$\pm$0.000} & 0.924\,{\scriptsize$\pm$0.000} \\
\bottomrule
\end{tabular}
\end{table*}

\begin{table}[t]
\centering
\caption{AUROC scores for both the video frame-count sweep and the rigorously re-encoded RodEpil cross-validation folds. For the sake of clarity, we have broken out the specific RodEpil AUROC results into their own dedicated section immediately below the frame-count data.}
\label{tab:aurocsweep}
\setlength{\tabcolsep}{6pt}
\renewcommand{\arraystretch}{1.15}
\begin{tabular}{@{}c ccc@{}}
\toprule
\textbf{Frames} & \textbf{Detection} & \textbf{3-class} & \textbf{5-class} \\
\midrule
1 & 0.615 & 0.599 & 0.634 \\
4 & 0.983 & 0.950 & 0.926 \\
8 & 0.989 & 0.957 & 0.919 \\
16 & 0.993 & 0.962 & 0.946 \\
32 & \textbf{0.994} & \textbf{0.971} & \textbf{0.951} \\
64 & \underline{0.994} & \underline{0.971} & \underline{0.950} \\
\bottomrule
\end{tabular}

\vspace{3mm}
\resizebox{\columnwidth}{!}{%
\begin{tabular}{@{}l ccccc c@{}}
\toprule
\textbf{RodEpil (re-enc.)} & \textbf{F0} & \textbf{F1} & \textbf{F2} & \textbf{F3} & \textbf{F4} & \textbf{Mean} \\
\midrule
AUROC & 0.999 & 1.000 & 1.000 & 1.000 & 0.998 & \textbf{0.999} \\
\bottomrule
\end{tabular}}
\end{table}

\subsection{Robustness of the cross-modal ordering}
\label{sec:extraresults:robustness}

While our primary analysis relies on a GRU model to mirror standard rodent-EEG setups~\cite{ganguly2025}, our broader conclusions do not hinge on that specific architecture. To test this, we swapped in a high-performing Temporal Convolutional Network (TCN) to process the EEG data. As the diagnostic tasks grew progressively harder, the TCN posted macro-F1 scores of $0.978\pm0.001$ for basic detection, $0.767\pm0.007$ for 3-class grading, and $0.563\pm0.012$ for the 5-class grading task.

When we line these numbers up against our standard video model ($0.968$, $0.780$, and $0.657$, respectively), the exact same pattern emerges: the TCN pulls slightly ahead in basic detection, effectively ties during 3-class grading, and then completely loses pace during the fine-grained 5-class severity task. This perfectly mirrors the performance trajectory of the baseline GRU ($0.974$, $0.750$, and $0.560$). Ultimately, this confirms that our core takeaway that video fundamentally outperforms EEG for granular severity grading is a genuine reflection of the modalities themselves, not an artifact of an artificially weak electrical baseline. (Though whether high-density, multi-channel EEG arrays might eventually close this gap remains an open question).

The underlying metrics tell a perfectly consistent story. When tracking the performance gap between the video and EEG models, the macro-F1 difference swings sharply from a microscopic $-0.006$ deficit in basic detection to a decisive $+0.097$ advantage for video during 5-class grading. The MCC captures this exact same inversion, flipping from a $-0.013$ deficit to a $+0.103$ lead. Because the AUROC metric completely saturates on this dataset and loses its ability to meaningfully separate the networks (Sec.~\ref{sec:extraresults:auroc}), relying on macro-F1 and MCC provides the most honest, realistic picture of actual diagnostic capability.

\subsection{Subject-exclusive transfer}
\label{sec:extraresults:subjecttransfer}

While session-exclusive splits prevent temporal leakage between adjacent frames, they still allow models to memorize animal-specific traits within the training pool. To enforce a far more demanding and realistic evaluation of generalization, we conducted a subject-exclusive 5-fold cross-validation, ensuring networks were tested strictly on unseen subjects. Reassuringly, basic detection generalizes exceptionally well. The video model suffers a negligible $0.020$ drop in macro-F1 (sliding from $0.968$ to $0.948\pm0.016$), while the EEG model barely shifts ($0.974$ to $0.972\pm0.015$). Corresponding MCC scores remain similarly solid at $0.896\pm0.031$ and $0.944\pm0.030$, respectively.

However, forcing the models to grade specific severity levels on completely unfamiliar animals takes a steep toll across the board. On the 3-class grading task, the video model's macro-F1 slides to $0.720\pm0.028$, while EEG drops to $0.656\pm0.022$. Under the grueling 5-class evaluation, video drops to $0.475\pm0.024$ and EEG falls to $0.404\pm0.042$. While video maintains a $0.071$ advantage over EEG on this hardest task, its overall drop in performance ($0.182$) is actually larger than EEG's ($0.156$). Ultimately, this shows that video is not immune to cross-subject variation: while both modalities easily generalize basic detection to unseen animals, mapping fine-grained severity signatures onto novel targets remains a significant challenge for both.

\subsection{Long-tail behavior and weighting}
\label{sec:extraresults:longtail}

The 5-class severity grading task forces us to confront severe class imbalance alongside massive variations across individual subjects. Looking at the raw counts, the dataset is heavily skewed: we have $12{,}197$ baseline clips versus $1{,}435$ Stage\,2, $9{,}256$ Stage\,3, $1{,}050$ Stage\,4, and just $193$ Stage\,5 events. Compounding this challenge, the volume of data contributed by each animal varies wildly, ranging from a meager 11 clips to as many as $3{,}091$ (Table~\ref{tab:cohort}). To stop the network from simply ignoring rare classes to minimize overall error, we used inverse-frequency-weighted cross-entropy loss (keeping the average weight normalized to unity). In practice, this forces the model to treat a single, ultra-rare Stage\,5 seizure with roughly $60\times$ the mathematical weight of a normal baseline clip.

Applying this weighting scheme forces the network to learn representations for the rarest seizure categories. For context, a naive majority-class classifier achieves a macro-F1 of just $0.136$. In contrast, our weighted video network reaches an overall macro-F1 of $0.657$, maintaining viable performance even on data-starved classes ($0.362$ F1 on Stage\,4 and $0.415$ on Stage\,5). The EEG model, however, still struggles despite the weighting, managing only $0.202$ and $0.123$ F1 for these respective stages (Table~\ref{tab:perclass}).

While this strategy successfully prevents class collapse, we deliberately relied on standard inverse-frequency weighting rather than evaluating unweighted baselines, synthetic oversampling techniques~\cite{chawla2002smote}, or focal loss formulations~\cite{lin2017focal}. Ultimately, our findings demonstrate that straightforward class weighting is sufficient to preserve model sensitivity across rare stages, establishing a practical baseline without requiring complex long-tail optimization strategies.

\subsection{Where each modality fails, and why}
\label{sec:extraresults:failure}

\begin{figure*}[t]
\centering
\includegraphics[width=\textwidth]{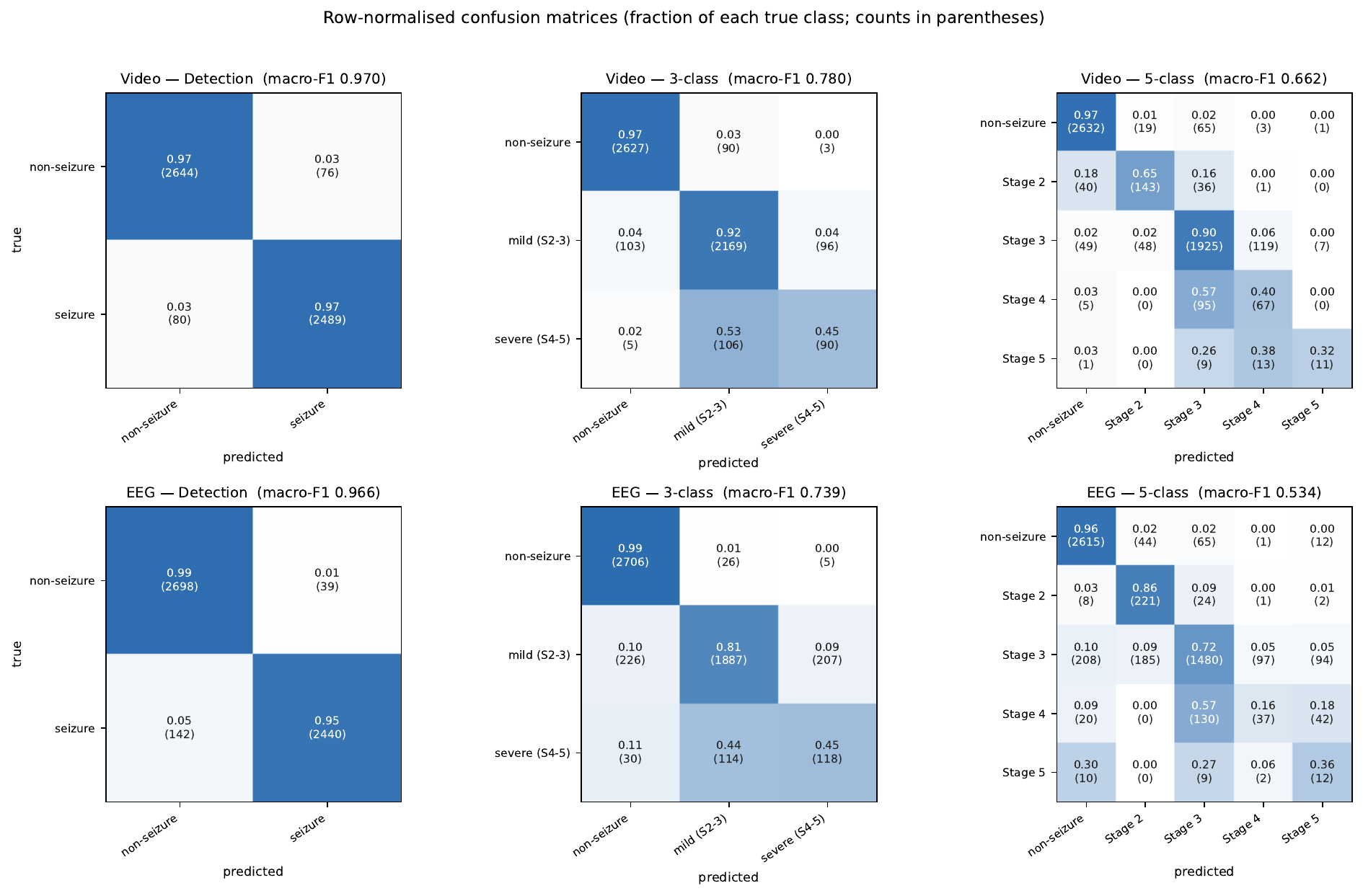}
\caption{Row-normalized confusion matrices comparing both modalities across all three evaluation tasks (drawn from a representative seed-42 diagnostic run; displaying true-class fractions with raw counts in parentheses). On the granular 5-class task, the video model's mistakes cluster tightly along the diagonal, meaning its errors are almost entirely limited to adjacent severity levels. In stark contrast, the EEG baseline not only misses active seizures completely, but also frequently commits erratic, non-adjacent grading errors.}
\label{fig:confusion}
\end{figure*}

\begin{figure*}[t]
\centering
\includegraphics[width=\textwidth]{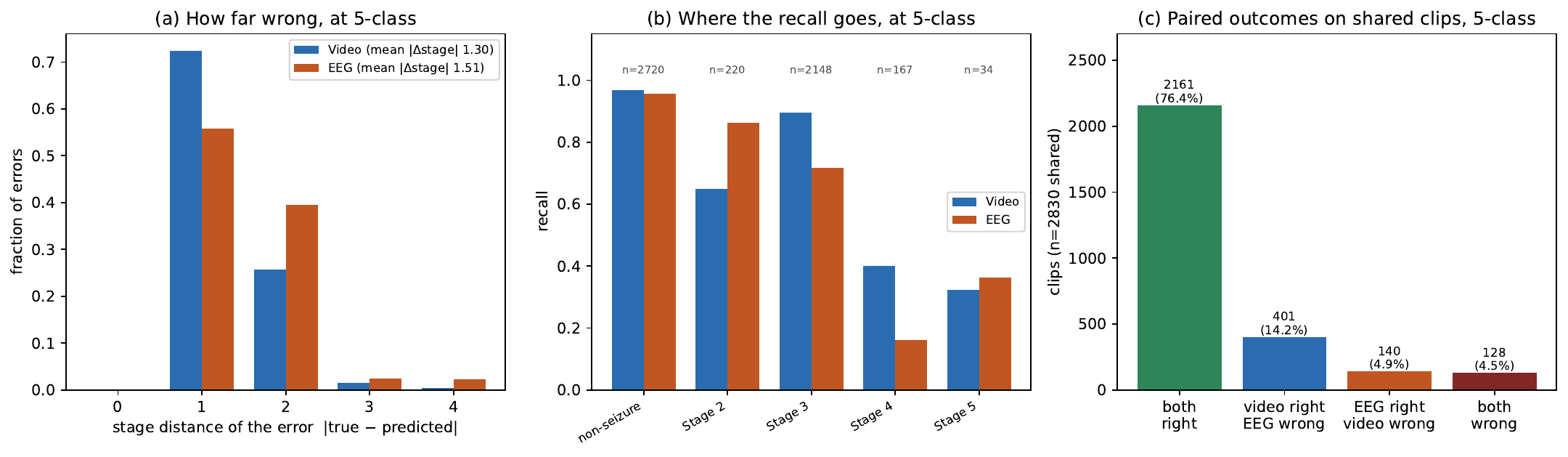}
\caption{Error diagnostics for the 5-class severity grading task. \emph{(a)} Numeric label-distance histogram. Because baseline non-seizure behavior falls completely outside the standard Racine scale, this panel remains purely descriptive and should not be treated as a strict ordinal metric. \emph{(b)} Per-class recall annotated with absolute support counts, distinctly highlighting the Stage\,2 performance reversal. \emph{(c)} Direct head-to-head outcomes across the strictly shared intersection of $2{,}830$ test clips evaluated by both modalities.}
\label{fig:failuremodes}
\end{figure*}

Table~\ref{tab:perclass} reveals that the video and EEG models exhibit strikingly distinct failure modes. Figure~\ref{fig:confusion} maps out these modal-specific confusion patterns, which we further illustrate with representative qualitative examples in Fig.~\ref{fig:failuremodes}.

\noindent\textbf{Missing a seizure entirely vs. misjudging its severity.}
The EEG model suffers a substantial drop in performance by failing to detect high-activity events entirely, missing 10.1\% of major Stage\,3 convulsions ($208$ out of $2{,}064$) compared to just 2.3\% for the video model ($49$ out of $2{,}148$). (We explicitly separate complete non-detections from severity grading errors, as baseline behavior lies outside the ordinal scale of seizure severity).

When grading detected events, the video network's misclassifications almost exclusively fall into adjacent severity levels such as mislabeling a Stage\,4 convulsion as Stage\,3 in 95 out of 167 error cases. In sharp contrast, the EEG model exhibits wide-ranging, non-adjacent errors, frequently confusing severe Stage\,5 episodes with milder Stage\,3 or Stage\,4 predictions. This confirms that video representations naturally capture the continuous physical progression of seizure behavior, whereas single-channel EEG signals struggle to resolve granular intensity without incurring erratic cross-stage confusions.

\noindent\textbf{The advantage flips at Stage\,2.}
Crucially, this dynamic completely reverses at Stage\,2. Because early-stage seizures manifest as subtle, low-motor behaviors, EEG recall spikes to $0.863$, whereas the video model lags at $0.650$. On identical paired evaluation clips, Stage\,2 stands out as the single category where EEG reliably outperforms video (Table~\ref{tab:paired}). This outcome reinforces our central hypothesis: vision-based models dominate only when seizures produce overt physical movement.

Evaluating joint predictions across paired clips highlights the deeply complementary strengths of both modalities. Specifically, the video model correctly identifies seizures missed by EEG across 14.2\% of the dataset, whereas EEG uniquely detects active events on 4.9\% of clips. Because their error distributions are largely non-overlapping, relying on either sensor in isolation introduces distinct diagnostic blind spots. Ultimately, capturing the full biological continuum of seizure dynamics requires multi-modal fusion rather than reliance on a single signal source.

\begin{table}[tb]
\centering
\caption{Direct head-to-head evaluation across the $2{,}830$ held-out test clips shared between both modalities for 5-class severity grading. To ensure a strictly fair comparison, this analysis is restricted to the exact intersection of clips evaluated by both models. Per-class rows highlight ``solo victories'' instances where precisely one model correctly predicted the severity stage while the other failed. Notably, Stage\,2 represents the sole category where the EEG model outperforms the video network in solo correct predictions.}
\label{tab:paired}
\setlength{\tabcolsep}{5pt}
\renewcommand{\arraystretch}{1.15}
\begin{tabular}{@{}l c c c@{}}
\toprule
\textbf{True class} & \textbf{n} & \textbf{video only} & \textbf{EEG only} \\
\midrule
non-seizure & 1488 & 73  & 44 \\
Stage\,2    &  192 & 7   & \textbf{44} \\
Stage\,3    & 1062 & \textbf{286} & 43 \\
Stage\,4    &   74 & 27  & 8  \\
Stage\,5    &   14 & 8   & 1  \\
\midrule
\multicolumn{4}{@{}l@{}}{\footnotesize overall: both correct $2161$ ($76.4\%$),
video only $401$ ($14.2\%$),}\\
\multicolumn{4}{@{}l@{}}{\footnotesize EEG only $140$ ($4.9\%$), both wrong $128$ ($4.5\%$)}\\
\bottomrule
\end{tabular}
\end{table}

\noindent\textbf{Motion-energy analysis.}
Using mean absolute frame difference as a straightforward proxy for physical motion (Table~\ref{tab:motion}) reveals that EEG recall for Stage\,3 events improves significantly during high-activity intervals ($0.777$ \vs\ $0.537$). (Crucially, because motion energy is derived exclusively from visual pixels rather than electrical telemetry, this correlation remains purely descriptive; it cannot serve as a direct causal explanation for EEG failures).

\begin{table}[tb]
\centering
\caption{Recall evaluated across the low- and high-motion halves of each severity class, partitioned strictly at the class-specific median motion energy ($900$ clips sampled from the paired evaluation set). We define motion energy as the mean absolute frame difference across $24$ evenly spaced $96\times96$ grayscale frames. Because this metric is computed over the entire duration of the clip including its $10$\,s pre- and post-event buffers it serves as a coarse but objective proxy for the overall physical intensity of the seizure.}
\label{tab:motion}
\setlength{\tabcolsep}{4pt}
\renewcommand{\arraystretch}{1.15}
\begin{tabular}{@{}l c c cc cc@{}}
\toprule
 & & & \multicolumn{2}{c}{\textbf{Video recall}} & \multicolumn{2}{c}{\textbf{EEG recall}} \\
\cmidrule(lr){4-5}\cmidrule(lr){6-7}
\textbf{True class} & \textbf{n} & \textbf{med.} & \textbf{low} & \textbf{high} & \textbf{low} & \textbf{high} \\
\midrule
non-seizure & 466 & 1.53 & 0.983 & 0.966 & 0.906 & 1.000 \\
Stage\,2    &  58 & 1.57 & 0.655 & 0.586 & 0.966 & 0.862 \\
Stage\,3    & 350 & 1.78 & 0.891 & 0.943 & 0.537 & \textbf{0.777} \\
Stage\,4    &  21 & 2.17 & 0.182 & \textbf{0.700} & 0.182 & 0.000 \\
\bottomrule
\end{tabular}
\end{table}

\noindent\textbf{Motion alone cannot explain the Stage\,2 anomaly.}
Crucially, basic physical movement completely fails to explain what is happening at Stage\,2. The video model's recall remains practically identical regardless of whether the overall motion is low ($0.655$) or high ($0.586$). Whatever factors are actually driving the performance gap between video and EEG during these subtle, early-stage seizures, raw motion magnitude alone is clearly not the answer.

\section{Extended Interpretation}
\label{sec:discussionfull}

Ultimately, our empirical findings point to a strict division of labor rather than a single, universally superior diagnostic modality. While EEG holds its ground during basic binary detection, video completely dominates the moment seizures escalate into observable physical movement. This perfectly reflects the underlying reality of the task: because seizure severity (Stages\,2 to 5) is fundamentally defined by motor behavior, video measures the clinical ground truth directly. EEG, on the other hand, is forced to infer these physical events through indirect electrographic proxies. Therefore, the resulting performance gap does not necessarily imply that EEG lacks severity information entirely, but rather that video benefits from a direct sensory alignment with the behavioral grading scale.

The performance inversion at Stage\,2 perfectly illustrates this reality. Because the physical movements during these early-stage seizures are so subtle, EEG easily dominates recall ($0.863$ \vs\ $0.650$). However, once the seizures escalate into violent Stage\,3 convulsions, severe movement artifacts likely overwhelm the telemetry equipment, causing the EEG model to completely miss 10.1\% of these major events (compared to a mere 2.3\% for video). Yet, as our low/high motion splits demonstrated, raw kinetic magnitude alone cannot fully account for this breakdown. Instead, these diagnostic blind spots are driven by a complex interplay of animal pose, event duration, electrical artifact masking, and the underlying electrophysiological signatures.

Operationally, visual detection proves remarkably robust across drastically different model sizes. The fact that a lightweight 4M-parameter X3D-M network can practically match the performance of a massive 33M-parameter R(2+1)D model proves a critical point: dense temporal tracking matters far more than sheer parameter count. This creates a clear path toward highly efficient, edge-deployable monitoring systems (though evaluating real-world streaming latency and continuous false-positive rates remains strictly outside the scope of this study).

Regardless of video's dominance in granular scaling, EEG remains non-negotiable for diagnosing ``silent'' internal electrographic anomalies lacking overt behavioral symptoms. Their wildly non-overlapping failure regimes practically mandate future multimodal architectures, specifically utilizing quality-aware attention protocols to suppress video during occlusions and suppress EEG during intense physical movement artifacts. Our unimodal framework establishes the exact baselines required to validate these future integrated systems.

\section{Limitations and Future Work in Detail}
\label{sec:openfull}

Sec.~\ref{sec:open} broadly defined five structural limitations; here, we brutally interrogate their impact against the empirical data.

\noindent\textbf{(1) Cross-subject and cross-cohort generalization.}
While subject-exclusive partitioning guarantees the network never memorizes an animal, every subject still stems from a localized, single-laboratory cohort. Under strict isolation, binary detection transfers phenomenally ($0.968 \to 0.948\pm0.016$), but granular grading collapses entirely ($0.657 \to 0.475\pm0.024$). This verifies within-cohort stability but provides zero guarantees regarding cross-laboratory hardware swaps, differing chemical induction protocols, or alternate rodent genetics. Massive multi-cohort benchmarking represents the ultimate deployment barrier.

\noindent\textbf{(2) Fine-grained and ordinal severity.}
Robust classification of extreme severity tiers remains a significant open challenge. Rare Stage\,4 and Stage\,5 episodes suffer from disproportionately high error rates, highlighting an acute need for targeted dataset expansion. Furthermore, because standard cross-entropy loss penalizes all misclassifications equally, it completely fails to exploit the inherent $\mathrm{S2}\prec\mathrm{S3}\prec\mathrm{S4}\prec\mathrm{S5}$ progression of seizure development. Future architectures must incorporate ordinal objectives to explicitly penalize non-adjacent errors based on this biological hierarchy. While Section~\ref{sec:problem:ordinal} outlines a theoretical conditional framework to address this, it still requires rigorous empirical validation. Similarly, our inverse-frequency weighting serves merely as a practical baseline; comprehensive comparisons against focal loss formulations~\cite{lin2017focal} or synthetic oversampling techniques~\cite{chawla2002smote} are necessary. Finally, because classical Racine scales were originally developed across varying chemical induction models~\cite{racine1972,vanerum2019racine}, distinguishing genuine network failures from underlying human annotation ambiguity will ultimately require rigorous, consensus-driven expert mapping.

\noindent\textbf{(3) EEG capacity and multimodal fusion.}
Our electrical baseline relies exclusively on a single recording channel, artificially constraining the spatial resolution available to the network. It remains entirely plausible that a dense, multi-electrode montage could capture richer spatial dynamics, potentially narrowing or even completely eliminating the grading advantage currently held by the visual model. Furthermore, while this study deliberately evaluated both modalities in strict isolation to establish their independent capacities, their sharply contrasting failure regimes clearly point toward integration. Future architectures must prioritize quality-aware video-EEG fusion protocols capable of dynamically suppressing modality-specific artifacts and resolving diagnostic conflicts, ultimately combining the complementary strengths of both sensors~\cite{frontiers_fusion2026, multimodal_review2026}.

\noindent\textbf{(4) Shared benchmarks.}
Although our uniform re-encoding of the RodEpil repository~\cite{rodepil2025} reduces identified encoding differences but does not rule out residual shortcuts, the complete lack of synchronized telemetry in this dataset renders direct multimodal comparisons impossible. To advance the field, there is an urgent need for a large-scale, open-source dataset that pairs precisely synchronized video with dense electrophysiology. Establishing guaranteed subject-exclusive data partitions alongside rigorous, fine-grained severity annotations will be absolutely essential for standardizing benchmarking efforts across competing network architectures.

\noindent\textbf{(5) Continuous and resource-constrained deployment.}
Our evaluation relies on pre-segmented clips, shielding the models from the complexities of continuous data streams. Real-world continuous monitoring requires low-latency edge inference, automated event boundary detection, and near-zero false-positive rates over extended recording periods. While lightweight backbones like X3D-M suggest that edge deployment is feasible and clinical frameworks such as vEpiNet~\cite{vepinet2024} demonstrate broader viability the practical engineering challenges of continuous, low-latency animal monitoring remain completely unaddressed and will require substantial future development.

\noindent\textbf{Annotation provenance and intertwined bias.}
Ultimately, because the ground-truth labels originate from a simultaneous human review of both data streams, neither modality is evaluated against a strictly independent baseline. In practice, the EEG telemetry inherently drove the temporal event boundaries, while the visual phenotype dictated the assigned severity grade. Resolving these intertwined biases will require future datasets to employ segregated, modality-specific expert annotations validated by rigorous inter-rater consensus. Only by doing so can we definitively decouple the true performance of the networks from the mechanics of the annotation process itself.

\end{document}